\documentclass[nohyperref]{article}

\usepackage{hyperref}

\usepackage[accepted]{include/icml2022}
\renewcommand\cite{\citep}

\def\final{0}

\newcommand\newcontent[1]{\textcolor{black}{#1}}
\newcommand\rockft{\usefont{T1}{fvs}{m}{n}}
\newcommand\rockfb{\usefont{T1}{fvs}{b}{n}}
\newcommand{\ROCKY}{{\rockft ROCK}}
\newcommand{\ROCKYTITLE}{{\rockfb ROCK}}
\newcommand{\ROCKYEMOJI}{\raisebox{-2pt}{\includegraphics[width=0.9em]{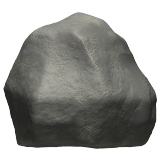}}}

\usepackage[utf8]{inputenc} 
\usepackage[T1]{fontenc}    
\usepackage{url}            
\usepackage{booktabs}       
\usepackage{amsfonts}       
\usepackage{amsmath,amsthm,amssymb,enumerate}
\usepackage{thm-restate}
\usepackage{nicefrac}       
\usepackage{microtype}      
\usepackage{epsfig}
\usepackage{color}
\usepackage{ifthen}
\usepackage{float}
\usepackage{alltt}
\usepackage{colortbl}
\usepackage{nicefrac}  
\usepackage{newlfont} 
\usepackage{floatflt}
\usepackage{wrapfig}
\usepackage{multirow,makecell}
\usepackage[capitalize,noabbrev]{cleveref}
\usepackage[export]{adjustbox}
\usepackage{mathtools, cuted}
\usepackage{dsfont}
\usepackage{graphicx}
\usepackage{appendix}
\usepackage{subcaption}
\usepackage{tcolorbox}
\usepackage{inconsolata}
\usepackage{verbatim}
\usepackage{enumitem}

\usepackage{tikz}
\usetikzlibrary{positioning,decorations,arrows,shapes,arrows.meta,calc,scopes,shadows,fadings}
\usepackage{tikzscale}

\usepackage[figuresright]{rotating}

\newcommand*\circled[1]{\tikz[baseline=(char.base)]{\node[shape=circle,
draw,inner sep=0.5pt,align=center] (char) {#1};}}

\newcommand*\uarrow[1]{\tikz[baseline=(char.base)]{->,draw,\draw[anchor=base,line width=1mm] (0,0) -- (0,0.1);}}

\newcommand{\ace}{\operatorname{ACE}}
\newcommand{\cau}{\operatorname{\mathsf{Cau}}}
\newcommand{\pa}{\operatorname{pa}}

\newcommand{\avg}{\operatorname{avg}}

\newcommand{\abs}[1]{\left\lvert #1 \right\rvert}

\newcommand{\norm}[1]{\left\lVert #1 \right\rVert}

\newcommand{\supp}{\operatorname{supp}}

\renewcommand{\Pr}{\mathbb{P}}
\newcommand{\Exp}{\mathbb{E}}
\newcommand{\Var}{\mathbf{Var}}

\newcommand{\distas}{\mathbin{\sim}}

\newcommand{\ind}[1]{\mathds{1}{\left\{#1\right\}}}
\newcommand{\eps}{\epsilon}

\renewcommand{\phi}{\varphi}

\newcommand{\ld}{\lambda}

\newcommand{\sR}{\mathbb R}

\newcommand{\calA}{{\cal A}}

\newcommand{\calE}{{\cal E}}

\newcommand{\calX}{{\cal X}}

\newcommand{\sfA}{{\sf A}}
\newcommand{\sfB}{{\sf B}}

\newcommand{\sfE}{{\sf E}}

\newcommand{\sfN}{{\sf N}}

\newcommand{\sfS}{{\sf S}}

\newcommand{\sfX}{{\sf X}}

\renewcommand{\o}{{\boldsymbol o}}

\newcommand{\q}{{\boldsymbol q}}

\renewcommand{\t}{{\boldsymbol t}}

\newcommand{\x}{{\boldsymbol x}}

\newcommand{\red}[1]{\textcolor{red}{#1}}
\newcommand{\blue}[1]{\textcolor{blue}{#1}}

\newcommand{\reviewer}[3]{
  \expandafter\newcommand\csname #1\endcsname[1]{
    \ifthenelse{\equal{\final}{1}} {
      \textcolor{#3}{}
    } {
    \textcolor{#3}{[\textsf{\footnotesize \textbf{#2:} ##1]}}
    }
  }
}

\ifdefined\email
\else
\newcommand{\email}[1]{\url{#1}}
\fi

\newtheorem{fact}{Fact}[section]
\newtheorem{lemma}[fact]{Lemma}
\newtheorem{theorem}[fact]{Theorem}
\newtheorem{definition}[fact]{Definition}

\newtheorem{assumption}[fact]{Assumption}
\theoremstyle{definition}
\newtheorem{problem}[fact]{Problem}
\newenvironment{namedthm}[1]{\newline
    \textbf{Theorem #1.}\itshape
}{
\newline}
\newtheorem{example}[fact]{Example}

\crefname{theorem}{theorem}{theorem}
\Crefname{theorem}{Theorem}{Theorems}
\crefname{assumption}{assumption}{assumption}
\Crefname{assumption}{Assumption}{Assumptions}
\crefname{lemma}{lemma}{lemma}
\Crefname{lemma}{Lemma}{Lemmas}
\crefname{definition}{definition}{definitions}
\Crefname{definition}{Definition}{Definitions}
\crefname{corollary}{corollary}{corollaries}
\Crefname{corollary}{Corollary}{Corollaries}
\crefname{proposition}{proposition}{propositions}
\Crefname{proposition}{Proposition}{Propositions}
\crefname{claim}{claim}{claims}
\Crefname{claim}{Claims}{Claims}
\crefname{problem}{problem}{problems}
\Crefname{problem}{Problem}{Problems}
\crefname{solution}{solution}{solutions}
\Crefname{solution}{Solution}{Solutions}
\crefname{proof}{proof}{proofs}
\Crefname{proof}{Proof}{Proofs}
\crefname{proofof}{proof}{proofs}
\Crefname{proofof}{Proof}{Proofs}

\newcommand{\bbm}{\begin{bmatrix}}
\newcommand{\ebm}{\end{bmatrix}}
\newcommand{\be}{\begin{equation}}
\newcommand{\ee}{\end{equation}}
\newcommand{\beg}{\begin{example}}
\newcommand{\eeg}{\end{example}}
\newcommand{\ben}{\begin{equation*}}
\newcommand{\een}{\end{equation*}}
\newcommand{\bea}{\begin{aligned}}
\newcommand{\eea}{\end{aligned}}
\newcommand{\ba}{\begin{equation}\begin{aligned}}
\newcommand{\ea}{\end{aligned}\end{equation}}
\newcommand{\bd}{\begin{definition}}
\newcommand{\ed}{\end{definition}}
\newcommand{\bt}{\begin{theorem}}
\newcommand{\et}{\end{theorem}}
\newcommand{\bnt}[1]{\begin{namedthm}{#1}}
\newcommand{\ent}{\end{namedthm}}
\newcommand{\blm}{\begin{lemma}}
\newcommand{\elm}{\end{lemma}}
\newcommand{\bp}{\begin{proof}}
\newcommand{\ep}{\end{proof}}
\newcommand{\bpb}{\begin{problem}}
\newcommand{\epb}{\end{problem}}
\newcommand{\benum}{\begin{enumerate}}
\newcommand{\eenum}{\end{enumerate}}
\newcommand{\bitem}{\begin{itemize}}
\newcommand{\eitem}{\end{itemize}}
\definecolor{firebrick}{RGB}{178,34,34}

\DeclareOldFontCommand{\rm}{\normalfont\rmfamily}{\mathrm}
\DeclareOldFontCommand{\sf}{\normalfont\sffamily}{\mathsf}
\DeclareOldFontCommand{\tt}{\normalfont\ttfamily}{\mathtt}
\DeclareOldFontCommand{\bf}{\normalfont\bfseries}{\mathbf}
\DeclareOldFontCommand{\it}{\normalfont\itshape}{\mathit}
\DeclareOldFontCommand{\sl}{\normalfont\slshape}{\@nomath\sl}
\DeclareOldFontCommand{\sc}{\normalfont\scshape}{\@nomath\sc}

\makeatletter
\newread\pin@file
\newcounter{pinlineno}
\newcommand\pin@accu{}
\newcommand\pin@ext{pintmp}
\newcommand*\partialinput [3] {%
  \IfFileExists{#3}{%
    \openin\pin@file #3
    \setcounter{pinlineno}{1}
    \@whilenum\value{pinlineno}<#1 \do{%
      \read\pin@file to\pin@line
      \stepcounter{pinlineno}%
    }
    \addtocounter{pinlineno}{-1}
    \let\pin@accu\empty
    \begingroup
    \endlinechar\newlinechar
    \@whilenum\value{pinlineno}<#2 \do{%
      \readline\pin@file to\pin@line
      \edef\pin@accu{\pin@accu\pin@line}%
      \stepcounter{pinlineno}%
    }
    \closein\pin@file
    \expandafter\endgroup
    \scantokens\expandafter{\pin@accu}%
  }{%
    \errmessage{File `#3' doesn't exist!}%
  }%
}
\makeatother

\newcommand{\abgood}[1]{{\bf\boldmath \blue{#1}}}
\newcommand{\abpoor}[1]{{\bf \red{#1}}}

\reviewer{dan}{dan}{Orange}
\reviewer{weijie}{Weijie}{BrickRed}
\reviewer{hongming}{Hongming}{JungleGreen}
\reviewer{jiayao}{Jiayao}{MidnightBlue}
\reviewer{TODO}{TODO}{Violet}

\begin{document}

\twocolumn[

\icmltitle{\ROCKYTITLE{\ROCKYEMOJI}: Causal Inference Principles\\for Reasoning about Commonsense Causality}
\icmlsetsymbol{equal}{*}
\begin{icmlauthorlist}
\icmlauthor{Jiayao Zhang}{cog,stats}
\icmlauthor{Hongming Zhang}{cog,tencent}
\icmlauthor{Weijie J. Su}{stats}
\icmlauthor{Dan Roth}{cog,aws}
\end{icmlauthorlist}

\icmlaffiliation{cog}{Cognitive Computation Group, University of Pennsylvania, USA.}
\icmlaffiliation{stats}{Department of Statistics and Data Science, University of Pennsylvania, USA.}
\icmlaffiliation{tencent}{Tencent AI Lab Seattle, USA.}
\icmlaffiliation{aws}{Amazon AWS AI Labs, USA}

\icmlcorrespondingauthor{}{\texttt{\{zjiayao,hzhangal,danroth,suw\}@upenn.edu}}

\icmlkeywords{natural language processing,commonsense reasoning,potential-outcomes framework}

\vskip 0.3in
]
\printAffiliationsAndNotice{}



\begin{abstract}
    Commonsense causality reasoning (CCR) aims at identifying
    plausible causes and effects
    in natural language descriptions that are \emph{deemed reasonable by
    an average person}. Although being of great academic and
    practical interest, this problem is still shadowed by
    the lack of
    a well-posed theoretical framework; existing
    work usually relies on deep language models 
    wholeheartedly,
    and is potentially susceptible to \emph{confounding co-occurrences}.
    Motivated by classical causal principles,
    we articulate the central question of CCR and
    draw parallels between human subjects in observational
    studies and natural languages to adopt CCR to the potential-outcomes
    framework which, to the best of our knowledge, is the first such attempt for commonsense
    tasks.
    We propose a novel framework,
    \ROCKY, to \textbf{R}eason \textbf{O}(A)bout
    \textbf{C}ommonsense \textbf{K}(C)ausality,
    which utilizes temporal signals as incidental supervision,
    and balances confounding effects using
    \emph{temporal propensities} that are
    analogous to propensity scores.
    \ROCKY{} is
    modular and zero-shot,
    and demonstrates good CCR capabilities.
\end{abstract}

\section{Introduction} \label{sec:introduction}

    Commonsense causality reasoning (CCR) is an important yet
    non-trivial task in natural language processing (NLP)
    that
    exerts broad industrial and societal impacts
    \cite{kuipers1984commonsense,COPA,GLUCOSE,sap2020commonsense}.
    We articulate this task as 
\begin{quote}
{\emph{reasoning about cause-and-effect relationships between events in natural language descriptions that are deemed reasonable by an average person.}}
\end{quote}

\begin{figure}
    \centering
    \begin{adjustbox}{width=\columnwidth}
\tikzset{%
  >={Latex[width=2mm,length=2mm]},
    base/.style = {rectangle, rounded corners, draw=black,
                           minimum width=3cm,
                           minimum height=1.4cm,
                           inner sep=1pt,
                           inner xsep=1pt,
                        align=left,
                        text width=8cm,
			font=\LARGE,
                        },
    event_cause/.style = {base, fill=ForestGreen!30, minimum width=5cm,text width=6cm},
    event_covariate/.style = {base, fill=RawSienna!30,minimum width=7cm,text width=7cm},
    event_effect/.style = {base, fill=NavyBlue!20},
    event_interv/.style = {base, fill=Goldenrod!30,minimum width=8cm,text width=8cm},
    marker_dot/.style = {circle,fill,inner sep=3pt},
}
\begin{tikzpicture}{node distance=3cm,
    every node/.style={fill=white}, align=center}
    \def\t{0.056} 

    \draw[->,line width=2] (-3,0) -- (12,0) node[right] {\LARGE Time};

    \draw[thick,line width=0.5mm] (6.4, 0) node[marker_dot]{} to[out=135,in=-45,looseness=0.5]++ (-6.5, 0.6) node[event_cause,above,align=center] (E1) {$\sfE_1$: {\tt Alice entered a restaurant.}};

    \draw[thick, line width=0.5mm] (9.4, 0) node[marker_dot]{} to[out=90,in=-90,looseness=0.5]++ (0.5, 0.6) node[event_effect,above,align=center] (E2) {$\sfE_2$: {\tt Alice ordered a pizza.}};

    \draw[thick,line width=0.5mm] (3.2, 0) node[marker_dot]{} to[out=-135,in=45,looseness=0.5]++ (-3, -0.6) node[event_covariate,below,align=center] (X1) {$\sfX_1$: {\tt Alice felt hungry.}};

    \draw[thick,line width=0.5mm] (6.4, 0) node[marker_dot]{} to[out=-135,in=45,looseness=0.5]++ (3, -0.6) node[event_interv,below,align=center] (A1) {$\sfA_1$: {\tt Alice opened a food-delivery app.} };

    \draw[->,thick,line width=0.5mm] (E1.east) -- node[above,align=center,text width=2cm] {{\LARGE \emph{causes?}}} (E2.west);

\end{tikzpicture}
    \end{adjustbox}
    \caption{\textbf{An Example of CCR:} \emph{does $\sfE_1$ cause $\sfE_2$?} The temporal order $\sfE_1 \prec \sfE_2$ does not necessitate causation due to confounding co-occurrences (e.g., $\sfX_1$). Since when \emph{conditioning on} $\sfX_1$,
    a \emph{comparable} intervention
    $\sfA_1$ of $\sfE_1$ also precedes $\sfE_2$, the effect from $\sfE_1$
    to $\sfE_2$ shrinks.} \label{fig:example}
\end{figure}
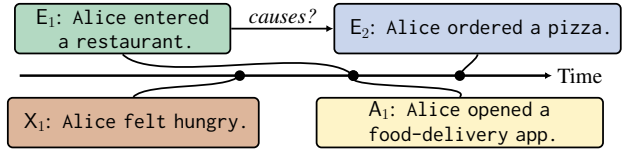

    This definition naturally excludes
    questions that are beyond commonsense knowledge, such as
    those scientific in nature
    (e.g., does a surgery procedure reduce mortality?). Instead, it
    accommodates
    causal queries within the reach of an ordinary reasonable person.
    {As a concrete instantiation, 
    we consider the problem of defining
    and estimating the strength of causation
    from one given event, $\sfE_1$, to another,
    $\sfE_2$.}
    For example, in \Cref{fig:example},
    is Alice's ``entering a restaurant'' ($\sfE_1$) a plausible cause
    for her ``ordering a pizza'' ($\sfE_2$)? Although the
    precedence
    from $\sfE_1$ to $\sfE_2$ is logical, it might
    be less a ``cause'' compared with Alice's ``feeling hungry'' ($\sfX_1$).

    Temporality 
    informs
    causation, but it is still
    unclear how to account for confounding
    co-occurrences {(such as $\sfX_1$ in \Cref{fig:example})}.
    Motivated by causal inference principles (\Cref{sec:background}),
    we formulate CCR as estimating the \emph{change} in the likelihood
    of $\sfE_2$'s occurrence
    due to intervening $\sfE_1$
    (denoted by $\neg \sfE_1$):
    \ba \label{eq:centralquestion}
	\Delta = \Pr \left( \sfE_1 \prec \sfE_2 \right)
	-
	\Pr \left( \neg \sfE_1 \prec \sfE_2 \right)
    \ea
    where $\Pr(\cdot)$ can
    be estimated by pretrained language models (LMs)
    e.g., via masked language modeling
    (see \Cref{sec:inference}
    for implementation details).
    {The estimand $\Delta$
    measures the \emph{average treatment effect} (ATE): its magnitude
    signifies the strength of the effect and its sign informs the direction. For example, when $\Delta$ is close to $-1$, $\sfE_1$ has a strong effect on $\sfE_2$ towards making $\sfE_2$ less prone to occurring.}
    If the occurrences of $\sfE_1$ and $\neg \sfE_1$
    on any unit are purely
    random, a direct estimation
    of the temporal probabilities in \Cref{eq:centralquestion} suffices;
    however, due to confounding
    co-occurrences (e.g., $\sfX_1$), one needs to \emph{balance}
    the covariates (events that precede $\sfE_1$) to eliminate
    potential spurious correlations.
    We propose \emph{temporal propensity}, a surrogate
    propensity score that can be used to balance
    the covariates (\Cref{sec:causal}).
    We show in \Cref{sec:simulations}
    that \emph{although temporality is essential
    for CCR, it is vulnerable to spurious
    correlations without being properly balanced}.

    \textbf{Contributions.} We articulate
    CCR from a completely new perspective
    using causal inference principles,
    and
    our contributions include (i) a novel commonsense causality framework;
    (ii) mitigating confounding co-occurrences
    by matching temporal propensities;
    (iii) a modular pipeline for zero-shot CCR with demonstrated
    effectiveness.


\section{Background} \label{sec:background}
    The problem of reasoning about causal relationships, and differentiating
    them from innocuous associations has been contemplated and studied
    extensively in human populations research spanning 
    clinical trials,
    epidemiology,
    political and social sciences, economics,
    and many more~\cite{fisher1958cancer,cochran1965planning,rosenbaum2002observational,imbens2015causal}
    among which causal practitioners usually base
    on
    the potential outcomes framework 
    \cite[also known as the Rubin causal model, see][]{neyman1923application,rubin1974estimating,holland1986statistics}, graphical and structural equation models \cite{robins1986graph,pearl1995scm,heckman2005rejoinder,peters2017causal}, and
    Granger causality \cite{granger1969causal}.

    With the recent celebrated empirical success of language models on various NLP tasks, especially
    transformers \cite{devlin2019bert,radford2019gpt}, there is an increasing
    interest in the NLP community on drawing
    causal inference using textual data.
    The majority of these works treat textual
    data as either covariates or study units \cite{keith2020text,feder2021causal} on which
    causal queries are formed (e.g., does taking a medicine affect
    recovery, which are recorded
    in textual medical records?). On the other hand, CCR
    with natural language
    descriptions struggles to fit in a causal inference
    framework:
    \emph{
    textual data in this case are just vehicles conveying semantic meanings,
    not to be taken at face value,
    }
    hence it is difficult
    to draw the parallel between causal inference that requires
    a clear definition of study units, treatments, and outcomes.

\subsection{Existing Approaches}

    Existing works related to CCR are usually grouped
    under the umbrella term of commonsense reasoning \cite{rashkin2018event2mind,ning2019joint,sap2020commonsense} or causal event detection
    \cite{ogorman2016riched}.
    Some of the notable progress usually comes from leveraging explicit causal
    cues/links (tokens such as ``due to'') and use conditional
    probabilities to measure ``causality'' \cite{Chang2004CausalRE,DoChRo11,luo2016commonsense}; leveraging large-scale pre-trained LMs via augmenting training datasets, designing
    training procedures, or loss functions
    \cite{sap2019social,shwartz2020selftalk,tamborrino2020pretraining,Zhang_2021_CVPR,staliunaite2021advcause}.
    
    There are several works that are relevant to ours,
    yet different in various ways:
    Granger causality, which measures association, is used by
    \citet{kang2017detecting} to detect event causes-and-effects;
    \citet{bessiere2020order} studies events as point-processes, in a way arguably closer to association;
    \citet{gerstenberg2021csm} uses a simulation model
    to reason physical causation.
    To the best of our knowledge,
    we are the first one to adopt a causal perspective in solving
    CCR.

\subsection{Challenges of CCR}

    Many existing CCR methods (mostly supervised)
    are based on ingenious designs
    and creative LM engineering.
    Theoretical
    justifications, however, are sometimes
    desirable, as only then
    do we know how
    general these methods can be.
    Indeed, recent studies reveal
    that several supervised
    models may have exploited
    certain artifacts in datasets 
    to ace the evaluations
    \cite{kavumba2019bcopa,han2021weak}.

    This dilemma of constructing a well-founded
    theoretical framework
    versus engineering models to achieve
    excellent empirical performances
    is not surprising, perhaps, given that the challenges of CCR from causal perspectives are
    not trivial at all:
    what is the study unit, treatment, and outcome in this case? What does it mean
    to ``intervene'', or ``manipulate'' the treatment? Is treatment \emph{stable},
    or is it desirable to consider multiple versions of it?

\begin{figure*}
    \centering
    \begin{adjustbox}{width=\textwidth}
\tikzset{%
  >={Latex[width=2mm,length=2mm]},
    base/.style = {rectangle, rounded corners, draw=black,
                           minimum width=6cm,
                           minimum height=0.8cm,
                           inner sep=2pt,
                           inner xsep=10pt,
                           node distance=0.3em and 3cm,
                        align=left,
                        text width=6cm,
                           font=\sffamily},
    event_cause/.style = {base, fill=ForestGreen!30, minimum width=7cm,text width=9cm,node distance=2cm and 3cm},
    event_covariate/.style = {base, fill=RawSienna!30,node distance=0.7em and 3cm},
    event_effect/.style = {base, fill=NavyBlue!20,text width=9cm},
    event_interv/.style = {base, fill=Goldenrod!30,minimum width=7cm,text width=7cm,node distance=0.7em and 3cm},
    block_lm/.style={rectangle,draw=black,inner sep=3pt,thick,rounded corners=0,text opacity=1,fill opacity=0.5,font=\ttfamily},
    block_prompt/.style={rectangle,draw=gray,fill=gray!5,inner sep=2pt,thick,rounded corners=0,text opacity=1,fill opacity=0.5,font=\ttfamily},
    block_cov/.style={rectangle,thick,rounded corners=7,blend mode=overlay,opacity=0.5},
    wide_arrow/.style={line width=1.5mm,-triangle 60,postaction={draw, line width=3mm,shorten >=4mm, -}},
    num_marker/.style={circle,draw=black,thick,fill=white,align=center},
}
\pgfdeclarelayer{background}
\pgfsetlayers{background,main}
\begin{tikzpicture}{node distance=3cm,
    every node/.style={fill=white}, align=center}

    \node (E1)  [event_cause] at (0,0)  {\Large $\sfE_1$: {\tt Alice walked into a restaurant.}};
    \node[block_lm,below=3.8cm of E1,align=center,fill opacity=0.9] (box_delta) {{\Large $\hat{\Delta}_p=
	    P(\textcolor{ForestGreen}{\sfE_1} \prec \textcolor{NavyBlue}{\sfE_2})
	- \avg_{\textcolor{Orange}{\sfA\in\calA'}}
	    P (\textcolor{Orange}{\sfA} \prec \textcolor{NavyBlue}{\sfE_2})$}};
    \node (X1) [event_covariate, below left=-1em and 3cm of E1] {$\sfX_1$: {\tt Alice felt hungry.}};
    \node (X2) [event_covariate, below=of X1] {$\sfX_2$: {\tt Alice went to a gym.}};
    \node (X3) [event_covariate, below=of X2] {$\sfX_3$: {\tt Alice turned on a light.}};
    \node (X4) [event_covariate, below=of X3] {$\sfX_4$: {\tt Alice took a train.}};
    \node (X5) [event_covariate, below=of X4] {$\sfX_5$: {\tt Bob went to a theater.}};
    \node (X6) [event_covariate, below=of X5] {$\cdots$};

    \node (A1) [event_interv, below right=-1em and 3cm of E1] {$\sfA_1$: {\tt Alice walked into a school.} };
    \node (A2) [event_interv, below=of A1] {$\cdots$ };
    \node (A3) [event_interv, below=of A2,fill=Orange!50] {$\sfA_1'$: {\tt Alice left from a restaurant.} };
    \node (A4) [event_interv, below=of A3,fill=Orange!50] {$\sfA_2'$: {\tt Alice opened a food-delivery app.} };
    \node (A5) [event_interv, below=of A4,fill=Orange!50] {$\sfA_3'$: {\tt Bob walked into a restaurant.} };
    \node (A6) [event_interv, below=of A5,fill=Orange!50] {$\cdots$};


    \begin{pgfonlayer}{background}
	\draw[block_cov,fill=RawSienna!20] ($(X1.north west)+(-0.2,0.6)$) rectangle ($(X6.south east)+(0.2,-0.6)$);
	\draw[block_cov,fill=Goldenrod!20] ($(A1.north west)+(-0.3,0.6)$) rectangle ($(A6.south east)+(0.3,-0.7)$);
	\draw[block_cov,fill=Orange!80] ($(A3.north west)+(-0.2,0.2)$) rectangle ($(A6.south east)+(0.2,-0.6)$);

    \end{pgfonlayer}

    \node[text width=5cm,above=-0.1em of X1] {$\calX:$ Sampled Covariates};
    \node[text width=5cm,above=-0.1em of A1] {$\calA:$ Generated Interventions};
    \node[text width=5cm,below=0.1em of A6] {$\calA':$ Matched Interventions};

    \draw[wide_arrow,draw=Goldenrod!50] ($(E1.south)+(1,-0.01)$) |- ++(0,-1.3cm) 
    -- ++(6.5cm,0);


    \draw[wide_arrow,draw=RawSienna!50] ($(E1.south)+(-1,-0.01)$) |- ++(0,-1.3cm) 
    -- ++(-6.6cm,0);

    \draw[wide_arrow,draw=Orange!50] ($(X6.east)+(0.2,-0.4)$) -- 
    ($(A6.west)+(-0.3,-0.4)$);

    \draw[wide_arrow,draw=Orange!50] ($(A3.west)+(-0.3,0)$) -| ($(box_delta.north)+(1.5,0.01)$);
    \draw[wide_arrow,draw=ForestGreen!50] ($(E1.south)+(-0.5,-0.01)$) --  ($(box_delta.north)+(-0.5,0.01)$);




    \node[num_marker,fill=RawSienna!20] at (-6.6, -1) (num1) {{\LARGE $1$}};
    \node[num_marker,fill=Goldenrod!20] at (1.8, -1) (num2) {{\LARGE $2$}};
    \node[num_marker,fill=Orange!20] at (-6.6, -5.6) (num3) {{\LARGE $3$}};
    \node[num_marker,fill=NavyBlue!20] at (-6.6, -3.6) (num4) {{\LARGE $4$}};

    \node (E2)  [event_effect, below=1.8cm of E1] {\Large $\sfE_2$: {\tt Alice ordered a pizza.}};

    \draw[wide_arrow,draw=NavyBlue!50] ($(E2.south)+(0.5,-0.01)$) --  ($(box_delta.north)+(0.5,0.01)$);

    \node[block_lm,right=0.2 cm of num1,align=center,text width=4cm,fill=RawSienna!50,fill opacity=0.5] {{\Large Sampling Prior Events}};
    \node[block_lm,right=0.2 cm of num2,align=center,text width=4cm,fill=Goldenrod!50,fill opacity=0.5] {{\Large Generating Interventions}};
    \node[block_lm,right=0.2 cm of num3,align=center,text width=8cm,fill=Orange!50,fill opacity=0.5] {{\Large Matching Temporal Propensities}};
    \node[block_lm,right=0.2 cm of num4,align=center,text width=4cm,fill=NavyBlue!50,fill opacity=0.5] {{\Large Estimating $\Delta$}};

\end{tikzpicture}

    \end{adjustbox}
    \caption{\textbf{Illustration of the \ROCKY{} framework.}
	\emph{Does $\sfE_1$ cause $\sfE_2$?} To answer this query,
    \circled{1} the event sampler samples a set
    of covariates $\calX$ of events $\sfX_k$ that occur
    preceding $\sfE_1$.
    \circled{2} The intervention generator generates a set $\calA$ of
    interventions $\sfA_k$ on $\sfE_1$.
    \circled{3} A subset $\calA' \subset\calA$ of interventions
    is selected  whose
    temporal propensities
    $q(\x; \sfA)$ is close to that of $\sfE_1$, $q(\x;\sfE_1)$
    (\Cref{eq:estimatingeqn}).
    \circled{4} The temporal predictor uses
    $\calA'$ to estimate $\Delta$.
    } \label{fig:framework}
\end{figure*}
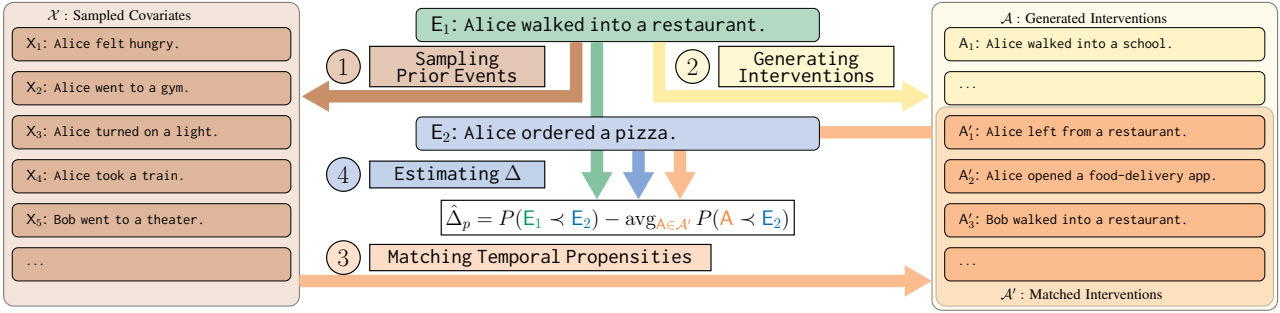

    \subsection{Principles of the \ROCKY{} Framework}
    In this paper, we attempt to 
    address 
    these questions using, among several causal principles,
    the following two that are intuitive and directly
    appeal to human nature
    \cite[see e.g.,][]{russell1912cause,bunge1979causal}: (1) \textbf{Precedence does not imply causation}, which warns us \emph{post-hoc}
    fallacies; (2) \textbf{Causation implies precedence}, which informs us that the
    events must be compared
    with those that are
    \emph{in pari materia}
    \cite{mill1851logic,hill1965ea},
    or having \emph{balanced} 
    covariates \cite[also called ``pretreatments,'' by which
    we mean events that occur prior to $\sfE_1$, cf.~][]{rosenbaum1989optimal}.
    Our CCR formulation in terms of
    temporality has
    several benefits:
    (i) the intrinsic temporality of causal principles characterizes
    its central role in CCR;
    (ii) temporal signals bring about
    incidental supervision \cite{Roth17,ning2019joint};
    (iii) although being a non-trivial question \emph{per se}, reasoning
    temporality
    has witnessed decent progress lately, making it more accessible
    than directly detecting causal signals
     \cite{NingFeRo17,ning2018temp,NingSuRo19,zhou2020temporal,vashishtha2020temporal}.

\newcommand{\evts}{E}
\newcommand{\evtsets}{\calE}
\newcommand{\metric}{\operatorname{m}}

\section{The \ROCKY{} Framework} \label{sec:causal}
\paragraph{Notations.}
    We use sans-serif letters
    for events, uppercase serif letters for \emph{indicators}
    of whether the corresponding event occurs,\footnote{By ``occurs,'' we mean ``is observed.'' We treat them
    interchangeable in the rest of our paper.}
    and lowercase serif letters for
    the realizations of those indicators. For example, in \Cref{fig:example,fig:framework}, 
    $\sfE_1:$ ``{\tt Alice walked into a restaurant.},''    $E_1=\ind{\sfE_1\text{ occurs}}$ and
    $e_{1,i} = \ind{\sfE_1\text{ occurs to the }i\text{-th study unit}}$\footnote{Defined among other concepts in \Cref{sec:causal:units}.}.
    We view the occurrence of events as point
    processes $E(t)$ on
    $t\in\{0,1\}$ (e.g., present versus past).
    We use $\sfE_1 \succ \sfE_2$
    (resp.~$\prec$)
    to indicate that $\sfE_1$ occurs following (resp.~preceding) $\sfE_1$.
    We write $\Pr(\sfE_1\prec \sfE_2)
    =\Pr(E_1(0), E_2(1))$ and $\Pr(\sfE_2 \vert \sfE_1)
    =\Pr(E_2(1)\vert E_1(0))$. We write $P$ for estimates
    of $\Pr$,
    and
    omit measure-theoretic details\footnote{Let $\calE$ be the set
	of commonsense events we consider, the probability space
	we are working on is $(\calE \times \calE, \sigma(\calE \times \calE),\Pr)$.}.

	\paragraph{Overview of the \ROCKY{} framework.} We set
	the stage in this section and discuss implementation details  in \Cref{sec:inference}.
	Given $\sfE_1$ and
    $\sfE_2$, as shown in \Cref{fig:framework},
	\ROCKY{} samples
	the covariates set $\calX$
	and interventions set $\calA$,
	from which a matched subset
	$\calA'$ is selected via
	temporal propensities (\Cref{sec:causal:temporal_propensity}).
	The score $\Delta$ is then
	estimated by
	\Cref{eq:estimatingeqn}.

\subsection{The Central Question of CCR}
    Given two specific events $\sfE_1$ and $\sfE_2$, 
    as discussed in
    \Cref{sec:introduction}, we articulate
    CCR as the estimation of
    the change of temporal likelihood \emph{had} $\sfE_1$ been \emph{intervened}:
    \ba 
	\Delta = \Pr \left( \sfE_1 \prec \sfE_2 \right)
	-
	\Pr \left( \neg \sfE_1 \prec \sfE_2 \right)
    \ea
    which assumes values in $[-1,1]$
    and measures a form of the \emph{average treatment effect}.
    As these probabilities are eventually estimated
    from data,
    if there are confounding
    events $\sfX_k$ that always
    co-occur with $\sfE_1$ in the data itself,
    they will bias
    this estimation.
    To this end, it is
    necessary to first clear out several key notions associated
    with this causal query, and then properly define
    the intervention $\neg \sfE_1$.

\subsection{The Potential-Outcomes Framework} \label{sec:causal:units}
    One major challenge of framing a causal query for CCR is the ambiguity of the underlying mechanism.
    Unlike human populations research, where experiments and study units
    are obvious to define, it is not immediately clear what they are when
    faced with semantic meanings of languages \cite{zhang2022some}.
    Yet, we can draw parallels again between semantic meanings
    and human subjects via the following thought experiment:
    suppose each human subject keeps a journal detailing the complete
    timeline of her experiences since her conception, then we can treat each
    individual as a study unit where the temporal relations of events can be
    inferred from the journal. 
    
\newcontent{
    We can then formulate CCR in the language of the potential-outcomes framework. Given fixed events $\sfE_1$ and $\sfE_2$, let
    $\sfE_{1i}$ denote the event experienced by the $i$-th study unit
    at time $t=0$ when $\sfE_1$ is supposed to occur. Each unit
    is then associated with a treatment assignment $E_{1i} = \ind{\sfE_{1i}=\sfE_1}$,
    realizations of the covariates 
    $\x_i = (x_{ij})_{j=1}^N$ for
    $x_{ij} = \ind{\sfX_j\prec \sfE_{1i}}$, and
    two potential outcomes
    \ba
    \begin{cases}
        r_{0i} = \ind{\sfE_{1i,E_{1}=0} \prec \sfE_2}, \\
        r_{1i} = \ind{\sfE_{1i,E_{1}=1} \prec \sfE_2}.
    \end{cases}
    \ea
    Here $\sfE_{1i,E_{1}=1-E_{1i}}$ signifies the hypothetical scenario where this unit \emph{had} received the treatment
    assignment
    $1-E_{1i}$, when in fact it receives $E_{1i}$.
    Clearly, either $r_{0i}$ and 
    $r_{1i}$ can be observed, but not both. Our estimand
    $\Delta$ in \Cref{eq:centralquestion} is indeed
    the average treatment effect
    \ba
        \Delta = \Exp[r_1 - r_0] \equiv \Pr(\sfE_1 \prec \sfE_2)
        - \Pr(\neg \sfE_1 \prec \sfE_2).
    \ea
}%
    This identification naturally complies with the temporal nature of
    covariates \cite{rubin2005causal}, since by definition they are
    \emph{pretreatments} that take place \emph{before} the treatment.
    We shall now address the issue of intervention (manipulation).
    Generally speaking, events are complex, and therefore intervention in this case would be better interpreted in a broader sense
    than one particular type of
    manipulation such as negation.
    For example, with $\sfE_1$ being
    ``Alice walked into a restaurant,''
    suppose hypothetically, before $\sfE_1$,
    Alice did not walk into a restaurant ($\neg \sfE_1^1$),
    we can thus compare $\Pr(\sfE_1 \prec \sfE_2)$
    with $\Pr(\neg \sfE_1^1 \prec \sfE_2)$ to reason to what extent some event
    $\sfE_2$ can be viewed as the effect due to $\sfE_1$.
    However,
    this is
    not the complete picture: Alice may have walked into somewhere else such as
    a bar; she may have, instead of walked into, but left a restaurant;
    instead of Alice, perhaps it was Bob who walked into a restaurant.
    The temporal information between these events and $\sfE_2$
    are also likely to inform
    causation between $\sfE_1$ and $\sfE_2$,
    and they are no less interventions
    than negation. As such we interpret
    intervention in our framework in a broader sense, not necessarily only
    negation or the entailment of negations, but \emph{any events that leads
    to plausible states of counterfactuality}. We will denote
    all possible interventions of $\sfE_1$ as $\calA$.
    \paragraph{Remark.}
    The generally acknowledged
    \emph{stable unit treatment value assumption} \cite[SUTVA,][]{rubin1980bias} requires that
    for each unit there is only one version
    of the non-treatment.
    \newcontent{
    Nonetheless, as we noted in the above discussion,
    the nature of the CCR problem
    renders it tricky to define what
    constitutes the exact version of the
    non-treatment (what single event \emph{is}
    not having done something, exactly?). For
    ease of exposition, we allow
    interventions in \ROCKY{} to take on multiple versions.
    }
    

\subsection{Balancing Covariates}
    The direct estimation of $\Delta$ in
    \Cref{eq:centralquestion} is feasible only
    in an ideal world where those
    probabilities are estimated from randomized controlled trials (RCTs)
    such that the treatment ($E_1$)
    is assigned completely at random to study units.
    Due to confounding co-occurrences,
    events that precede $\sfE_1$
    need to be properly
    balanced \cite{mill1851logic,rosenbaum1983central,pearl2018book}.
    Taking again as an example $\sfE_1$: ``Alice walked into a restaurant,''
    and $\sfE_2$: ``Alice ordered a pizza,''. Suppose hypothetically,
    Alice's twin sister Alicia, who has the exact life experiences up to
    the point when $\sfE_1$ took place, but opted not to walk into a restaurant,
    but opened a food delivery app on her phone ($\neg \sfE_1$). Then
    we can reason that the cause-and-effect relationship from $\sfE_1$
    to $\sfE_2$ is perhaps not large. On the other hand, if we know
    another irrelevant person, say Bob, underwent $\neg \sfE_1$ and then
    $\sfE_2$, then perhaps we are not ready to give that conclusion since
    we do not know if Bob and Alice are comparable at the first place.
    This example illustrates the importance of adjusting or balancing pretreatments.
    \newcontent{
    As such, we may rewrite
    \Cref{eq:centralquestion} as conditional expectations among study units that are comparable, i.e.,
    \ba \label{eq:condiprob}
	\Exp_{\x} \left[ \Pr(\sfE_1 \prec \sfE_2 \vert \x)
		- \Pr(\neg \sfE_1 \prec \sfE_2 \vert \x) \right],
    \ea
    \emph{provided} that the treatment assignment
    is strongly ignorable with respect to $\x$,
    in the sense of the following assumption.
    \begin{assumption}[Strong Ignorability] \label{as:ignore}
    The potential outcomes $\{r_0,r_1\}$ are
    independent with the treatment assignment $E_1$ 
    conditioning on the covariates $\x$. 
\end{assumption}
}
    
    \paragraph{Remark.} (i) We should define $\x$
    as events preceding $\sfE_1$, but \emph{not} $\sfE_2$, which will potentially introduce
    posttreatment biases \cite{rosenbaum1984consequences}: if
    an $\sfX'$ that occurs between $\sfE_1$ and $\sfE_2$ is adjusted, $\Delta$ thus estimated
    quantifies the effect from $\sfE_1$ to $\sfE_2$ \emph{without} passing through $\sfX'$.
    (ii) Although $\x$ should be those that are correlated with $\sfE_1$, adjusting for un-correlated
    effects does not introduce biases.

\subsection{Matching Temporal Propensities} \label{sec:causal:temporal_propensity}

    There are several techniques for balancing covariates such as
    sub-classification, matched sampling, covariance adjustment, and via structural equations \cite{cochran1965planning,pearl1995scm}.
    \citet{rosenbaum1983central} showed
    that the propsensity score can be used for this purpose.
    The propensity score
    $p(\x) = \Pr(E_1(1) = 1\vert \x(0))$ is the probability of $\sfE_1$ occurring at time $1$ conditioning on the covariates being $\x$ at time $0$.

    To properly identify what events constitute the covariates
    set is essential for our CCR framework.
    In the best scenario,
    it should include the real cause(s), which is, however, exactly what CCR solves.
    To circumvent this
    circular dependency, we use large
    LMs to sample a large number of events preceding
    $\sfE_1$, which should provide a reasonable covariate
    set. In this case,
    directly computing $p(\x)$ is not feasible, as
    will be discussed in \Cref{sec:inference},
    instead, we propose to use a surrogate which we call \emph{temporal propsensities}:
    \ba \label{eq:temppropensity}
	q(\x)=q(\x;\sfE_1) = \left( \Pr(E_1(1) =1\vert x)\right)_{x\in\x}
    \ea
    with each coordinate measuring the conditional probability
    of the event $\sfE_1$ given an event in $\x$.
    Thus motivated,
    for some fixed threshold $\eps$ and $p\in\{1,2\}$,
    we will use following estimating equation for the
    $L_p$-balanced score, where $f(\sfE_1, \sfE_2)$
    is an estimate for $\Pr(\sfE_1 \prec \sfE_2)$:
    \ba \label{eq:estimatingeqn}
    \begin{cases}
	\hat{\Delta}_p =
	    f(\sfE_1,\sfE_2) - \frac{1}{\abs{\calA'}}
		\sum_{\sfA \in \calA'} f(\sfA,\sfE_2), \\
	\calA' \coloneqq
	\resizebox{0.7\columnwidth}{!}{%
	$\left\{\sfA \in \calA:
    \frac{1}{\abs{\calX}}\norm{q(\x; \sfA) - q(\x;\sfE_1)}_p \le \eps \right\}$}.
    \end{cases}
    \ea

\newcontent{
\subsection{Discussions on Temporal Propensity Matching}
Unfortunately, the estimator $\hat{\Delta}_p$ in \Cref{eq:estimatingeqn} 
is generally biased even if a perfect matching of temporal
propensity exists, because $q(\x)$ consists of
conditional probabilities on one-dimensional 
marginal distributions
instead of on the full joint distribution.
Quantifying this loss of information is a difficult problem by itself; here we outline a coarse
bound for illustration purposes. 
}
\begin{restatable}[Expected $L_2$ error under perfect matching]{proposition}{thmtpm}
    \label{thm:tpm}
    Write $r\coloneqq r_1-r_0$, then $\Delta = \Exp[r_1-r_0]\equiv \Exp[r]$.
    Define 
    \ba
        \varrho\coloneqq \sup_{\tau}\left\{
            \tau\le \abs{r - \Exp[r\vert q(\x)]} \text{ a.s.}  \right\}
            \in \{0,1\}.
    \ea
    The expected $L_2$ error of $\hat{\Delta}=\Exp[r\vert q(\x)]$ satisfies
    \ba
        \Exp[(\hat{\Delta}-\Delta)^2]
        \le 1 - \varrho^2.
    \ea
\end{restatable}
\newcontent{
The proof is due to the conditional variance decomposition and is given
in the Appendix. The parameter
$\varrho$ depends on the problem instance and quantifies
the level of dependence between the potential outcomes
$\{r_0,r_1\}$ and the treatment assignment $E_1$ when conditioned
on the covariates $\x$. 
Intuitively, the worst-case scenario $\varrho = 0$ is uncommon, since
this happens only if $r$ is a function of $q(\x)$.
When a large number
of \emph{diverse} covariates are sampled, $\varrho$ is unlikely to be 0.  We thus assume that $\varrho \gg 0$ and we can balance temporal propensities to produce
a reasonable estimate.
}




\section{Implementation of \ROCKY} \label{sec:inference}

    Having established a framework for CCR,
    we provide an exemplar implementation of \ROCKY{} in this section.
    Our purpose is to demonstrate the potential of the \ROCKY{}
    and we expect engineering
    efforts such as prompt design can bring further improvements.
    
    The core tool we shall use is (finetuned)
    pretrained deep LMs.
    With the sheer amount of training data (e.g.,
    over $800$GB for the Pile dataset,
    \citet{pile}),
    it is reasonable to assume
    that those models would imitate responses of an average reasonable person.
    On the other hand, it is hard for generation models
    (masked or open-ended) to parse information that
    are far from the mask tokens;
    instead, it is more feasible for LMs to sample
    summary statistics of the relationships between 
    a pair of events, which is one of the main
    motivations for using temporal propensities (\Cref{eq:temppropensity}).

\subsection{Components of \ROCKY}

    For practical purposes, we represent
    an event as a $3$-tuple $(\texttt{ARG0}, \texttt{V},\texttt{ARG1})$.
    \ROCKY{} takes two events $\sfE_1$
    and $\sfE_2$ as inputs, and returns an estimate $\hat{\Delta}$ for $\Delta$ according to \Cref{eq:estimatingeqn}. 
    It contains four components (cf.~\Cref{fig:framework}):
    an event sampler that samples a set $\calX$ of events that
    are likely to occur preceding $\sfE_1$; a
    temporal predictor whose output $f(\sfX_1,\sfX_2)$ given two input
    events $\sfX_1$ and $\sfX_2$ is an estimate of the temporal
    probability $\Pr(\sfX_1 \prec \sfX_2)$;
    an intervention generator that generates a set $\calA$ of events
    that are considered as interventions of the event $\sfE_1$;
    and finally a scorer that first forms the temporal propensity
    vectors $q(\x; \sfA) \in \sR^{\abs{\calX}}$
    for each sampled interventions $\sfA \in \calA$, then estimates
    $\Delta$ via \Cref{eq:estimatingeqn}.
    We next discuss in greater details our implementation of this pipeline.

\subsection{Implementation Details}

\paragraph{Event Sampling.} Given an event $\sfE_1$ (e.g., $\sfE_1:$ {\tt Alice walked into a restaurant.}), we construct the
    prompt by adding ``{\tt Before that,}'' to the sentence, forming
    ``{\tt Alice walked into a restaurant. Before that, }'' as the final
    prompt. We use the GPT-J model \cite{gpt-j} , which is pretrained on the Pile
    dataset \cite{pile} for open-ended text generation where
    we set max length of returned sequences to be $30$, temperature to be
    $0.9$. We sample $n=100$ events, cropping at the first
    stop token of the newly generated sentence to form $\calX$.

    \paragraph{Temporal Prediction.} Given two events $\sfE_1$ and
    $\sfE_2$, we use masked language modeling to predict their
    temporal relation by forming the prompt
    $\sfE_1 \text{ \tt   <MASK> } \sfE_2$ and collect the score
    $s_a(\sfE_1, \sfE_2)$ and $s_b(\sfE_1, \sfE_2)$
    for the tokens {\tt after} and {\tt before}. We then symmetrize
    the estimates to form $s(\sfE_1, \sfE_2)= \frac{1}{2}(s_a(\sfE_1, \sfE_2)
    + s_b(\sfE_2, \sfE_1))$. We can directly use $s(\sfE_1,\sfE_2)$
    for $f(\sfE_1,\sfE_2)$; we discuss possible normalizations of this
    score in \Cref{sec:simulations}.

    \paragraph{Temporality Fine-Tuning.}
    Directly using a pretrained LM as the temporal
    predictor is likely to suffer from
    low coverage, since the tokens {\tt before} and {\tt after}
    usually are not among the top-$k$ most probable tokens. We can increase $k$ but
    this does not heuristically justify if the outputted scores are
    meaningful.
    We thus use the New York Times (NYT) corpus which
    contains NYT articles from 1987 to 2007 \cite{Sandhaus08} to fine-tune an LM.
    Following the same procedure as \citet{zhou2020temporal}, we perform
    semantic role labeling (SRL) using AllenNLP's BERT SRL model \cite{Gardner2017AllenNLP} to identify sentences with a
    temporal argument ({\tt ARG-TMP})
    that starts with a temporal connective {\tt tmp}
    (either {\tt before} or {\tt after}). We then extract the verb and its two
    arguments (${\tt V, ARG0, ARG1}$) as well as this tuple from its
    temporal argument, thus forming an event pair $(\sfE_1, \sfE_2, \text{\tt tmp})$. We are able to extract $397174$
    such pairs and construct them into the fine-tuning dataset consisting of
    ``$\sfE_1 \text{ \tt tmp } \sfE_2$'' and ``$\sfE_2\text{ } \neg\text{\tt tmp } \sfE_2$''
    for each extracted pair,
    where $\neg\text{\tt tmp}$ is the reverse temporal
    connective (e.g., {\tt after} if {\tt tmp} is {\tt before}).
    We then fine-tune a pretrained RoBERTa model ({\tt RoBERTa-BASE}) using HuggingFace Transformers
    \cite{wolf2020hf} via mask language modeling with masking probability
    $p=0.1$ for each token. We choose a batch size of
    $500$ and a learning rate of $5\times 10^{-5}$, and train
    the model to convergence, which
    was around $135000$ iterations when the loss converges
    to $1.37$ from $2.02$.

    \paragraph{Intervention Generator.} Given event $\sfE_1$, the intervention
    generator generates a set $\calA$ of events that are considered
    as interventions of the event $\sfA$
    in the sense of \Cref{sec:causal:units},
    which includes manipulating
    \texttt{ARG0}, \texttt{V},
    and \texttt{ARG1} respectively.
    We achieve this goal by masking these components
    individually and filling in the masks using an LM.
    There are several
    existing works on generating interventions of sentences
    \cite{feder2021causal}, and we select {\tt PolyJuice} \cite{wu2021polyjuice} in our pipeline
    due to its robustness.
    {\tt PolyJuice} allows conditional generation via control codes
    such as {\tt negation}, {\tt lexical}, {\tt resemantic},
    {\tt quantifier}, {\tt insert}, {\tt restructure}, {\tt shuffle},
    and {\tt delete}, each corresponds to a different
    manner how
    the sentence is intervened. We drop the fluency-evaluation component
    of {\tt PolyJuice} as they will be
    evaluated by the temporal predictor.
    {We remark that
    in Figure~\ref{fig:example}, the
    intervention is not generated from 
    {\tt PolyJuice}. Nonetheless, such interventions can be produced by more elaborated LMs.}

    \paragraph{Score Estimation.} Given the interventions $\calA$
    and the sampled covariates $\calX$, we can use the temporal predictor
    to estimate $\Pr(\sfX \prec \sfA)$ for all $\sfX \in \calX$ and $\sfA\in\calA$.
    To obtain temporal propensities $q(\x; \sfA)$
    for all interventions, we need
    to estimate $\Pr(A=1 \vert X)$ for each $\sfX$ and $\sfA$.
    Since by our sampling method,
    $\sfX$
    \emph{occurs} preceding $\sfE_1$, 
    there is an implicit conditioning
    on $\sfE_1$,
    we may thus set
    $P(X(0)) = f(\sfX, \sfE_1)$ 
    and $P(X(0), A(1)) = f(\sfX, \sfA)$;
    we will discuss possible normalizations in \Cref{sec:exp:normalizations}.
    We then form temporal propensity vectors as
    (recall $X$ is the indicator corresponding to the
    event $\sfX$)
     \ba \label{eq:est:q}
        q(\x;\sfA) = \left( \frac{P(X(0))}{P(X(0), A(1))}\right)_{\sfX\in\calX}.
    \ea

\newcommand{\copa}{\textsc{COPA}}
\newcommand{\copadev}{\textsc{COPA-Dev}}
\newcommand{\copate}{\textsc{COPA-Test}}
\newcommand{\glt}{\textsc{GLUCOSE-D1}}
\section{Empirical Studies} \label{sec:simulations}
    We put the \ROCKY{} framework
    into  action\footnote{Code for the ROCK and for reproducing all results in this paper is available at \href{https://github.com/zjiayao/ccr_rock}{\tt github.com:zjiayao/ccr\_rock.git}.}, our findings reveal that \emph{although temporality
    is essential for CCR, without balancing covariates,
    it is prone to spurious correlations.}

\newcommand*{\belowrulesepcolor}[1]{%
  \noalign{%
    \kern-\belowrulesep
    \begingroup
      \color{#1}%
      \hrule height\belowrulesep
    \endgroup
  }%
}
\newcommand*{\aboverulesepcolor}[1]{%
  \noalign{%
    \begingroup
      \color{#1}%
      \hrule height\aboverulesep
    \endgroup
    \kern-\aboverulesep
  }%
}

\begin{table*}
    \footnotesize\centering
	\begin{tabular}{lp{1.5cm}p{1.7cm}p{1.7cm}p{1.7cm}p{1.7cm}p{1.7cm}}
        \toprule
	\, & Random & $\hat{\Delta}_1$ $\uparrow$& $\hat{\Delta}_2$ $\uparrow$& $\hat{\Delta}_{\sfE_1}$ $\uparrow$ & $\hat{\Delta}_{\calA}$ $\uparrow$ & $\hat{\Delta}_{\calX}$ $\uparrow$\\
	\, & Baseline & $L_1$-Balanced& $L_2$-Balanced &  Temporal & Unbalanced & Misspecified \\
        \midrule
	\copadev{} & $0.5\pm0.050$ & $0.6900$ & $\abgood{0.7000}$ & $0.5800$ & $0.5600$ & \abpoor{$0.5300$}  \\
	\copate{} & $0.5\pm0.022$ & $\abgood{0.5640}$ & $\abgood{0.5640}$ & \abpoor{$0.5200$} & $0.5400$ & $0.5240$  \\
	\glt{} & $0.5 \pm 0.040$ & $0.6645$ & $\abgood{0.6968}$ & $0.5677$ & \abpoor{$0.5742$} & $0.6581$  \\
	\midrule
	\belowrulesepcolor{Gray!20}
	\rowcolor{Gray!20}
	\copadev{} (-T) & $0.5\pm 0.050$ & $0.6200$ & \abgood{$0.6300$} & $0.5300$ & \abpoor{$0.4800$} & $0.5300$  \\
	\rowcolor{Gray!20}
	\copate{} (-T) & $0.5\pm 0.022$ & \abgood{$0.5800$} & $0.5740$ & \abpoor{$0.4540$} & $0.4600$ & $0.4860$ \\
	\rowcolor{Gray!20}
	\glt{} (-T) & $0.5\pm 0.040$ & $0.6065$ & \abgood{$0.6194$} & $0.5548$ & $0.4387$ & \abpoor{$0.3742$} \\
    \aboverulesepcolor{Gray!20}
    \bottomrule
    \end{tabular}
    \caption{\textbf{Best zero-shot results.} Shaded rows
    have temporal fine-tuning (T)
    disabled.
	(i) Estimators with temporal propensities
	balanced ($\hat{\Delta}_1$ and $\hat{\Delta}_2$)
	perform
	consistently better than the unbalanced and
    the temporal estimators.
    (ii) In general,
    without temporality fine-tuning  (``-T'', see \Cref{sec:inference}), the performances degrade.
}
    \label{tab:res:copadev}
\end{table*}

    \begin{figure*}
	\centering
	\includegraphics[width=\textwidth]{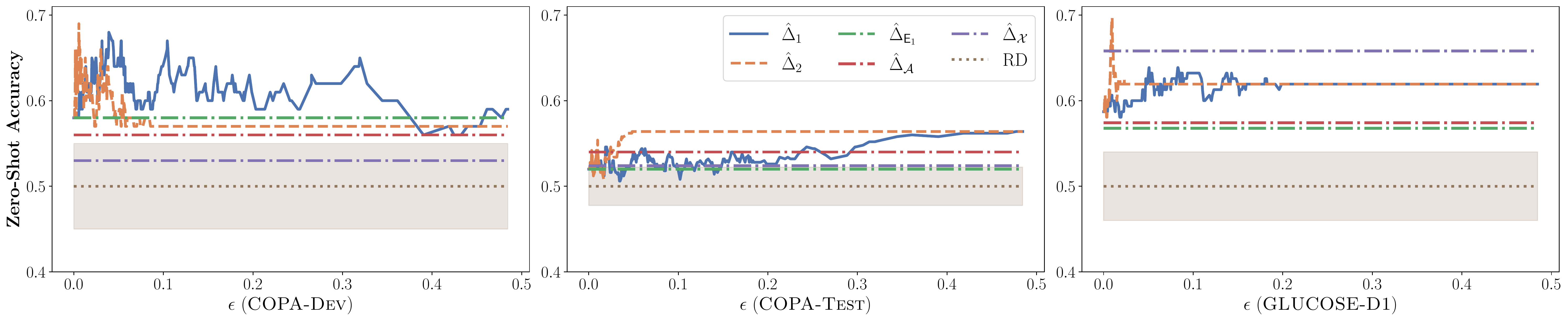}
	\caption{\textbf{Best zero-shot result vs $\eps$.} Balanced estimators significantly
	outperform un-balanced and other variants
	for both \copadev{} (left),
	\copate{} (middle) and \glt{} (right).} \label{fig:exp:res}
    \end{figure*}

\subsection{Setup and Details}

\paragraph{Evaluation Datasets.}

    We evaluate the \ROCKY{} framework on the Choice of Plausible Alternatives 
    dataset \cite[\copa,][]{COPA}
    and a self-constructed dataset of $153$
    instances using the first dimension (cause-and-effect) of GLUCOSE \cite[\glt,][]{GLUCOSE}.
    Each instance in \copa{} consists of a premise,
    two plausible choices, and a question type asking
    which choice is the choice (or effect) of the premise.
    When asking for cause, we set the premise as $\sfE_1$,
    and two choices as $\sfE_2$ respectively; otherwise
    we take the premise as $\sfE_2$ and two choices as $\sfE_1$ respectively.
    We choose the choice with the higher score.
    We evaluate the development set of $100$ instances (\copadev)
    and the test set of $500$ instances (\copate).
    To construct \glt, we take the test set
    and set the cause
    as premise, the effect and another
    candidate event as two
    choices then follow the same procedure.

\paragraph{Baseline Scores and Variants.}
    To test the validity and the effectiveness of
    \ROCKY{},
    We compare the adjusted
    score $\hat{\Delta}_p$ with several other reasonable scores
    that may be intuitive at first sight.
    \bitem
        \item $L_1$-balanced score $\hat{\Delta}_1$:
        set $p=1$ in~\eqref{eq:estimatingeqn}.
        \item $L_2$-balanced score $\hat{\Delta}_2$:
        set $p=2$ in~\eqref{eq:estimatingeqn}.
        \item Vanilla temporal score $\hat{\Delta}_{\sfE_1}=\Pr(\sfE_1 \prec \sfE_2)$.
        \item Unadjusted score $\hat{\Delta}_{\calA}$:
        set $\calA' = \calA$ in~\eqref{eq:estimatingeqn}.
        \item Misspecified score $\hat{\Delta}_{\calX}$:
        set $\calA'=\calX$ in~\eqref{eq:estimatingeqn}.
    \eitem
    Here the $L_p$-balanced scores
    are those balanced using temporal propensities
    with $L_p$ norm in \Cref{eq:estimatingeqn};
    the vanilla temporal score is perhaps the most
    straightforward one, which treats temporal precedence
    as causation;
    the unadjusted score is obtained without balancing the
    covariates; the misspecified score mistakes the covariates for interventions.
    All these three have intuitive explanations
    but are either insufficient for CCR or prone to spurious correlations.
    Note that $\lim_{\eps\downarrow 0}\hat{\Delta}_p=\hat{\Delta}_{\sfE_1}$ (when
    nothing is kept)
    and $\lim_{\eps\uparrow1}\hat{\Delta}_p=\hat{\Delta}_{\calA}$ 
    (when everything is kept).

\subsection{Design Choices and Normalizations} \label{sec:exp:normalizations}
    We discuss several design choices and
    normalizations that might stabilize estimation procedures.
    We give the complete ablation studies on all combinations of
    these choices in \Cref{sec:simulations:ablation}. We
    observe that
    although some of these normalization
    may benefit CCR on certain datasets, the improvements are
    \emph{marginal} compared with what
    temporal propensity matching brings.

	\paragraph{Direct Matching ($\mathbf{D}$).}
	    In \eqref{eq:est:q},
	    we directly match the vectors
	    of probabilities $\left( f(\sfA, \sfX) \right)_{\sfX\in\calX}$.
	    
	\paragraph{Temporality Pre-Filtering ($\mathbf{F}$).}
	    As the covariate sampler and temporal predictor are
	    two different LMs, a sampled covariate might not be
	    a preceding event judged by the temporal predictor.
	    We filter the covariates before matching temporal propensities
	    such that $f(\sfX, \sfE_1) > f(\sfE_1,\sfX)$.
	    
	\paragraph{Score Normalization ($\mathbf{S}$).}
	    In \Cref{sec:inference} we use $s(\sfE_1,\sfE_2)$ for $f(\sfE_1, \sfE_2)$,
	    we can also normalize it and form $f(\sfE_1, \sfE_2)$ through
	\ba
	\resizebox{0.8\columnwidth}{!}{%
	    $f(\sfE_1,\sfE_2) = \frac{s(\sfE_1, \sfE_2)}
	    {s(\sfE_1, \sfE_2) +s(\sfE_2, \sfE_1) +
		s(\sfE_1, \sfN) +s(\sfN, \sfE_1)}$},
	\ea
	where $\sfN$ represents the null event when no additional information
	is given, set as an empty string.
	
    \paragraph{Propensity Normalization ($\mathbf{Q}$).}
    In \Cref{eq:est:q}, we can also
    normalize the estimates first before
    forming the $q$ vectors via
    $P(X(0)) = f(\sfX, \sfE_1) / \sum_{\sfX'\in\calX} f(\sfX', \sfE_1)$
    and $P(X(0), A(1)) = f(\sfX, \sfA)/ \sum_{\sfX'\in\calX} f(\sfX', \sfA)$.
    

	\paragraph{Co-occurrence Stablization ($\mathbf{C}$).}
	    The fine-tuned temporal predictor may sometimes still fail
	    to cover the connectives.
	   We can stabilize $\Pr(\sfX \prec \sfA)$ by setting it to
	    $(P(A(0),X(1)) + P(X(0),A(1)))/2$.

    \paragraph{Estimand Normalization ($\mathbf{E}$).}
        We can normalize the 
        probability $\Pr(\sfA \prec \sfB)$
        in the estimand $\Delta$ 
        by dividing by $(P(A(0),B(1)) + P(B(0), A(1)))$.

\subsection{Results}
\subsubsection{A Concrete Example}
	We first examine a particular example when the vanilla temporal score
	$\hat{\Delta}_{\sfE_1}$ fails but $\hat{\Delta}_1$ does not.
	\begin{example}[Did $\sfE_1^{(1)}$
	or $\sfE_1^{(2)}$ cause $\sfE_2$?]
	{\small
	\begin{align*}
	    \sfE_1^{(1)}: &\quad \texttt{I was preparing to wash my hands.} \\
	    \sfE_1^{(2)}: &\quad \texttt{I was preparing to clean the bathroom.} \\
	    \sfE_2:&\quad \texttt{I put rubber gloves on.} \\
	    \sfA_{15}^{(1)}:&\quad \texttt{I was preparing to wash my feet.}\\
	    \sfA_5^{(2)}:&\quad \texttt{Kevin was preparing to clean the bathroom.}
	\end{align*}
	}
	\end{example}
	This is the $63$-rd instance in \copadev{}
	together a matched intervention
	($L_2$-balancing with optimal $\eps$)
	for each choice.
	The unadjusted scores are
	$\hat{\Delta}_{\calA}(\sfE_1^{(1)}, \sfE_2)\approx0.036$ and 
	$\hat{\Delta}_{\calA}(\sfE_1^{(2)}, \sfE_2)\approx0.035$
	while the $L_1$-balanced scores 
	are
	$\hat{\Delta}_{1}(\sfE_1^{(1)}, \sfE_2)\approx-0.010$ and 
	$\hat{\Delta}_{1}(\sfE_1^{(2)}, \sfE_2)\approx 0.002$. 
	The balanced score selects the correct
	choice ($\sfE_1^{(2)}$)
	with higher confidence. 
	More details 
	and full examples
	are given in the Appendix.
	We should comment that the scores
	$\hat{\Delta}_{1}$, $\hat{\Delta}_{\calX}$
	and $\hat{\Delta}_{\sfE_1}$ also select the correct
	answer on this instance; and there are instances where the balanced
	scores fail. Nonetheless, the performance of balanced scores dominates
	on average.

\subsubsection{Discussion}

    We show best zero-shot results
    over design choices (and over $\eps$) in \Cref{fig:exp:res} and \Cref{tab:res:copadev}.
    As \ROCKY{} tackles CCR from a completely
    new perspective,
    there are no real baselines to compare with;
    our goal is to demonstrate that
    \emph{the causal inference motivated method,
    temporal propensity matching,
    mitigates spurious correlations}
    by comparing balanced scores with unbalanced
    ones.
    We think this perspective would also benefit the
    NLP community at large for solving CCR and other related
    tasks.


\begin{figure*}[t!]
	\centering
    \includegraphics[width=\textwidth]{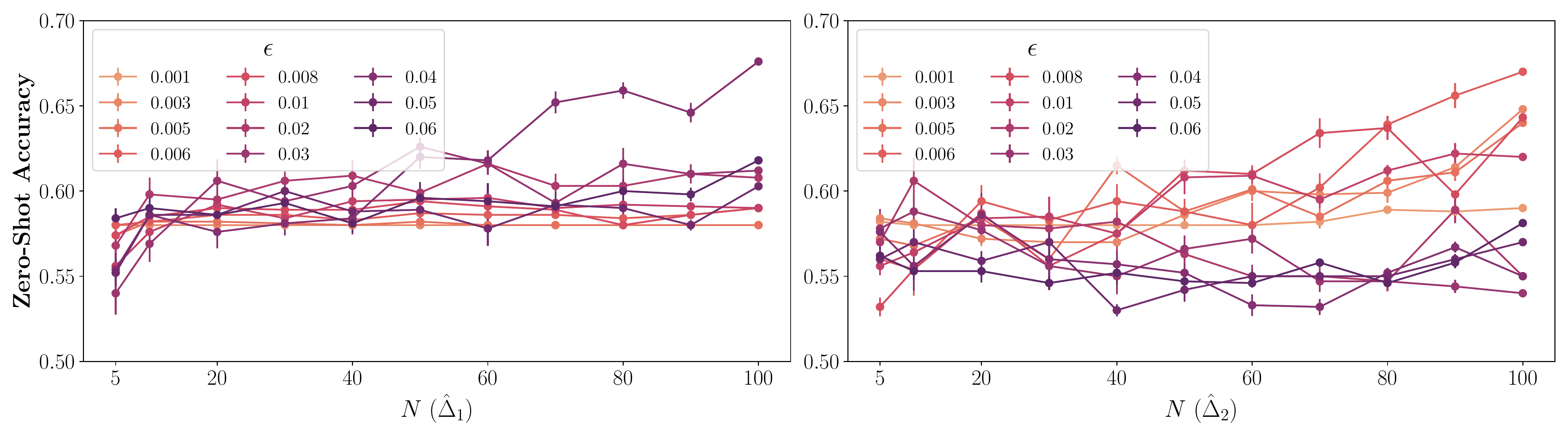}
	\caption{\textbf{Zero-shot result on \copadev{} vs covariate set size $N=\abs{\calX}$ with $95\%$-confidence bands.}
	In general, using a larger $N$
	improves performances for
	both $L_1$-balanced score ($\hat{\Delta}_1$, left)
	and $L_2$-balanced score ($\hat{\Delta}_2$, right).
	}
	\label{fig:exp:copaN}
\end{figure*}


    \textbf{Temporal propensity matching is effective.}
    In \Cref{tab:res:copadev} (unshaded rows), we observe that
    balanced scores have generally better performances on all
    datasets compared with the temporal estimator and the
    unadjusted estimator, implying that (i)
    temporality is important for CCR, yet they
    are susceptible to spurious correlations; (ii)
    balancing covariates
    via matching temporal propensities is effective.


    \textbf{Rules-of-thumb for choosing $\eps$.}
    The parameter $\eps$ controls the threshold of covariates selection
    and $p$ controls its geometry \cite[see e.g.,][]{htw15sparse}.
    Hinted by \Cref{fig:exp:res}, a general rule-of-thumb should be $\eps < 0.1$.
    \Cref{tab:res:besteps} shows optimal $\eps$ values when constrained to $[0,0.1]$,
    where all are global optimal except for \copate{} under $L_1$-balanced score
    (whose accuracy is $0.552$).
    Hence we recommend setting $\eps$ to be reasonably small $\eps$ such as
    within $(0.01, 0.1)$ when $p=1$ and relatively
    smaller such as
    $(0.005, 0.05)$ when $p=2$. The optimal value depends on the implementation details
    of \ROCKY{} components
    and domains of CCR to be performed, yet these choices should provide a good start.
   
\newcontent{
    \textbf{Comparison with existing methods.}
    The self-talk method \cite{shwartz2020selftalk}
	achieves $66\%$ on \copadev{} without external knowledge and $69\%$ when
	the CoMET-Net \cite{BRSMCC19} that contains commonsense knowledge is used.
	\citet{wei2022flan} reports $91\%$ on the training set of \copa{} by using instruction fine-tuning on related datasets.
	\citet{tamborrino2020pretraining} reports $80\%$ on \copate{}
	by ranking choices using an $n$-gram based scoring method.
	\ROCKY{} method outperforms
	self-talk but underperforms  \cite{wei2022flan,tamborrino2020pretraining}
	in its current form.
	Nonetheless, our method only requires temporal information provided by the \textbf{vanilla} LM without any task-specific fine-tuning,
	is more interpretable, and
	provides a prototype for adopting causal inference frameworks
	to natural language tasks.
}


\subsection{Ablation Studies} \label{sec:simulations:ablation}
    \textbf{Temporality Fine-Tuning.}
    Shaded rows in \Cref{tab:res:copadev} show that when we use 
    the pretrained \texttt{RoBERTa-BASE} without temporality fine-tuning (we increase $k$ to $30$),
    almost all estimators do not have decent performance.
    We conclude that (i) pretrained LMs
    usually have poor ``temporal
    awareness,'' and (ii) temporal fine-tuning
    helps LMs to extract temporal knowledge
    essential to CCR.

    \textbf{Covariate Set Size.}
    \Cref{fig:exp:copaN} depicts zero-shot
    results on \copate{} against the covariate set size $N = \abs{\calX}$
    together with $95\%$-confidence bands.
    Here we only enable score normalizations ($\sfN$) among all six normalizations.
    We observe that in general, increasing covariate
    set size improves performances if $\eps$
    is reasonable: if $\eps$ is too small, added covariates may have little
    impacts while they may introduce more noises if $\eps$ is too large.

    \textbf{Normalizations.}
    In \Cref{sec:exp:normalizations} we discussed six possible normalizations.
    We report the best performance when each
    normalization is removed in \Cref{tab:res:ablation}, where
    red marks the percentage decrease compared
    with the best result ({\bf D} and {\bf F} not shown as there is no change).
    Full ablations of all combinations of
    normalizations and more discussions are given
    in the Appendix. We observe that (i) certain normalizations benefit
    certain datasets; (ii) in general, improvements due to normalizations
    are only \emph{marginal}.


\begin{table}
    \footnotesize\centering
	\begin{tabular}{p{0.4cm}p{0.65cm}p{0.65cm}p{0.65cm}p{0.65cm}p{0.65cm}p{0.65cm}}
        \toprule
	\, & \multicolumn{2}{c}{\copadev} & \multicolumn{2}{c}{\copate} & \multicolumn{2}{c}{\glt} \\
	\, & $\hat{\Delta}_1$ $\uparrow$ & $\hat{\Delta}_2$  $\uparrow$ & $\hat{\Delta}_1$ $\uparrow$ & $\hat{\Delta}_2$  $\uparrow$ &$\hat{\Delta}_1$ $\uparrow$ & $\hat{\Delta}_2$  $\uparrow$ \\
        \midrule
	Best & $0.6900$ & $0.7000$ &$0.5640$ & $0.5640$ & $0.6645$ & $0.6968$ \\
        \midrule
    \textbf{-S} &  \abpoor{$0.01$} & \abpoor{$0.06$} & - & - & \abpoor{$0.08$} & \abpoor{$0.11$}  \\
    \textbf{-Q} &  \abpoor{$0.01$} & - & - & - & \abpoor{$0.03$} & -  \\
    \textbf{-C} &  - & - & \abpoor{$0.01$} & \abpoor{$0.01$} & \abpoor{$0.09$} & \abpoor{$0.13$}  \\
    \textbf{-E} &  \abpoor{$0.01$} & \abpoor{$0.01$} & - & - & \abpoor{$0.03$} & -  \\
    \bottomrule
    \end{tabular}
    \caption{\textbf{Single-component ablations on normalizations.} Marked in red
    are percentage decreases compared with the best result (i.e., computed as $(a-b)/a$).
}
    \label{tab:res:ablation}
    \vspace{-20pt}
\end{table}

\section{Discussions and Open Problems} \label{sec:conclusions}

We articulate the central question of CCR
and introduce \ROCKY,
a novel framework for zero-shot CCR,
which is the first attempt to incorporate causal inference frameworks
in commonsense reasoning.
\ROCKY{} sheds light on the CCR problem
from new perspectives 
that are arguably more well-founded
and demonstrates
great potential for zero-shot CCR 
as shown by empirical studies of various datasets and is on par with existing methods that leverages external causal knowledge on some datasets. 

There are
several possible avenues for future works.
(i) \textbf{Prompt engineering} for better
temporal predictors and event sampler
will likely benefit \ROCKY.
(ii) \textbf{Implicit events and reporting biases}
in training data are likely to bias the
LMs. How to
account for implicit events?
(iii) \textbf{Computing the exact propensity} requires
design novel methods to extract many-event temporal relationships and would further improve the performance.
{(iv) \textbf{Investigating implicit biases in the framework.} When the LM is sufficiently large and the pretraining dataset sufficiently diverse, the LM outputs should have reasonably well coverage and less bias due to undercoverage.}

\section*{Acknowledgements}
This work was supported in part by ONR Contract N00015-19-1-2620,
NSF through CCF-1934876, an Alfred Sloan Research Fellowship, and the Wharton Dean’s Research Fund. 
We would like to thank Bo Zhang, Rotem Dror, Ben Zhou, and Soham Dan for helpful discussions and feedback on this manuscript. We also thank Shuxiao Chen, Sihao Chen, Vered Schwartz, Haoyu Wang, Haoshu Xu, Diyi Yang, and Yachong Yang for stimulating discussions at various stages of this work.

%
%

    \bibliography{bib/ccg.bib,bib/cited.bib,bib/new_cited_srt.bib}
    \bibliographystyle{include/icml2022}
\clearpage


\setcounter{section}{1}\setcounter{equation}{0}
\numberwithin{equation}{section}
\counterwithin{figure}{section}
\counterwithin{table}{section}
\counterwithin{fact}{section}

\clearpage
\appendix

\section{Miscellaneous Proofs}
We first restate \Cref{thm:tpm} below.
\thmtpm*
\begin{proof}[Proof of \Cref{thm:tpm}]
Recall we write $r \coloneqq r_1 - r_0$, by the conditional variance
decomposition, we have
\ba
    \Var(r) =& \Exp\Var(r\vert q(\x)) + \Var\Exp[r\vert q(\x)] \\
            =& \Exp\left[
                \left(r - \Exp[r\vert q(\x)]\right)^2
            \right] \\
            &+ \Exp\left[
                \left(\Exp[r\vert q(\x)] - \Exp[r]\right)^2
            \right] \\
            \ge& \Exp\left[
                \left(\Exp[r\vert q(\x)] - \Exp[r]\right)^2
            \right] + \varrho^2.
\ea
Note that $\Var(r) \le 1$ since $r\in[0,1]$, we have
the expected $L_2$ error
\ba
    \Exp\left[
        \left(\Exp[r\vert q(\x)] - \Exp[r]\right)^2
    \right]
    \le 1 - \varrho^2.
\ea

\end{proof}

\section{Additional Experiment Details}
    
\subsection{Rule-of-Thumb for Choosing $\eps$}
    In \Cref{tab:res:besteps} we
    show the best $\eps$ values when
    constrained in $\eps\in[0,0.1]$.
Hence we recommend setting $\eps$ to be reasonably small $\eps$ such as
    within $(0.01, 0.1)$ when $p=1$ and relatively
    smaller such as
    $(0.005, 0.05)$ when $p=2$. The optimal value depends on the implementation details
    of \ROCKY{} components
    and domains of CCR to be performed, yet these choices should result in a good start.

\subsection{Further Discussions on Temporality Fine-Tuning}
    In \Cref{fig:exp:res}, we observe that,
    counterintuitively, without
    temporality fine-tuning, the best performances
    of balanced estimators ($0.58$) are higher than
    those with temporality fine-tuning ($0.564$).
    Although this gap is within one standard
    deviation of the random baseline ($0.022$)
    thus no statistically significant conclusions
    can be drawn, but  it might
    hint that pretrained LMs may have already
    been very aware of termpoality. Is this really the case?
    A closer look at the full ablation table
    to be introduced shortly in \Cref{tab:app:ablation:full} reveals that
    the stellar performance is attributed
    to one particular normalization,
    estimand normalization ({\bf E}),
    which was actually detrimental to another
    dataset (\glt). Hence we think this normliazation
    may favor certain dataset over others, thus
    we think it is not recommendable to include
    this normalization when dealing with a new
    dataset.

\subsection{Full Ablation on Normalizations}

    Recall in \Cref{sec:simulations:ablation} we
    discussed six possible normalizations
    that may stabilize the estimation procedure:
    
    \bitem 
	\item[(\textbf{D})] \textbf{Direct Matching:}
	    in \eqref{eq:est:q}, instead of forming
	    the temporal propensity vectors $\q$
	    using conditional probabilities,
	    we may directly match the vectors
	    of probabilities $\left( f(\sfA, \sfX) \right)_{\sfX\in\calX}$. This normalization 
	    is not well motivated but might be easier to
	    compute under certain circumstances, hence
	    we include it as a comparison.

	\item[(\textbf{F})] \textbf{Temporality Pre-Filtering:}
	    as the covariate sampler and temporal predictor are
	    two different LMs, a sampled covariate might not be
	    a preceding event judged by the temporal predictor.
	    Thus, we can filter the covariates $\calX$ before matching temporal propensities
	    such that we only keep covariates $\sfX\in\calX$ satisfying $f(\sfX, \sfE_1) > f(\sfS, \sfE_1)$.
	    
	\item[(\textbf{S})] \textbf{Score Normalization:}
	    in \Cref{sec:inference} we use $s(\sfE_1,\sfE_2)$ for $f(\sfE_1, \sfE_2)$.
	    We can also normalize it and form $f(\sfE_1, \sfE_2)$ through
	\ba
	    f(\sfE_1,\sfE_2) = 
	    \resizebox{0.4\columnwidth}{!}{%
	    $\frac{s(\sfE_1, \sfE_2)}
	    {s(\sfE_1, \sfE_2) +s(\sfE_2, \sfE_1) +
		s(\sfE_1, \sfN) +s(\sfN, \sfE_1)}$}
	\ea
	where $\sfN$ represents the null event when no additional information
	is given, set as an empty string. In practice,
	this normalization does not 
	differ much from the normalization
	\ba
	\resizebox{0.6\columnwidth}{!}{%
	    $f(\sfE_1,\sfE_2) = \frac{s(\sfE_1, \sfE_2)}
	    {s(\sfE_1, \sfE_2) +s(\sfE_2, \sfE_1)}$},
	\ea
	which does not involve $\sfN$. However, using
	$\sfN$ has the benefit of stabilizing the estimate
	$f(\cdot,\cdot)$ as in rare scenarios $s(\sfE_1, \sfE_2)$
	and $s(\sfE_2, \sfE_1)$ may both close to zero.
	
    \item[(\textbf{Q})] \textbf{Propensity Normalization:}
    in \Cref{eq:est:q}, we can also
    normalize the estimates first before
    forming the $q$ vectors via
    \ba 
        P(X(0)) &=  \resizebox{0.3\columnwidth}{!}{%
        $\frac{f(\sfX, \sfE_1)}{\sum_{\sfX'\in\calX} f(\sfX', \sfE_1)}$},\\
        P(X(0), A(1)) &= \resizebox{0.3\columnwidth}{!}{%
        $\frac{f(\sfX, \sfA)}{\sum_{\sfX'\in\calX} f(\sfX', \sfA)}$},
    \ea
    where we estimate $P(X(0))$ as the relative frequency
    of $X(0)$ among all possible events in $\calX$; and
    $P(X(0), A(1))$ among all possible $(\sfX, \sfA)$ pairs.

	\item[(\textbf{C})] \textbf{Co-Occurrence Stabilization:}
	on rare occasions, the fine-tuned temporal predictor may sometimes still fail to cover the connectives.
	   We can stabilize $\Pr(\sfX \prec \sfA)$ by setting it to $(P(A(0),X(1)) + P(X(0),A(1)))/2$.
    This in effect results in an alternative estimand based on
    co-occurrences of events (instead of precedence)
    and can be viewed
    as a weaker causation in CCR.

    \item[(\textbf{E})] \textbf{Estimand Normalization:}
        the score normalization ($\sfN$) takes
        place at temporal propensity matching.
        We can normalize the temporal
        probability $\Pr(\sfA \prec \sfB)$
        in the estimand $\Delta$ 
        by dividing $(P(A(0),B(1)) + P(B(0), A(1)))$,
        thus setting
    \ba
        \Pr(\sfA \prec \sfB)=
        \resizebox{0.4\columnwidth}{!}{%
        $\frac{P(A(0), B(1))}{P(A(0),B(1)) + P(B(0), A(1))}$}.
    \ea
    \eitem

    \begin{table}[t]
	\scriptsize\centering
	\begin{tabular}{lp{0.7cm}p{0.7cm}p{0.7cm}p{0.7cm}p{0.7cm}p{0.7cm}}
        \toprule
	\, & \multicolumn{2}{l}{\copadev} & \multicolumn{2}{l}{\copate} & \multicolumn{2}{l}{\glt} \\
	\, & $\hat{\Delta}_1$& $\hat{\Delta}_2$ & $\hat{\Delta}_1$ & $\hat{\Delta}_2$   &$\hat{\Delta}_1$  & $\hat{\Delta}_2$  \\
	\midrule
	 $\eps^*$  & 0.043067 & 0.006029 & 0.059232 & 0.048837 & 	0.046643 & 	0.009374\\
    \bottomrule
    \end{tabular}
    \caption{\textbf{Best choices of $\eps$ when $\eps < 0.1$.}} \label{tab:res:besteps}
\end{table}

\begin{table}
	\centering
	\resizebox{\columnwidth}{!}{%
	\begin{tabular}{lp{1.5cm}p{1.5cm}p{1.5cm}p{1.5cm}p{1.5cm}}
    \toprule
	\, & $\hat{\Delta}_1$& $\hat{\Delta}_2$ & $\hat{\Delta}_{\sfE_1}$ & $\hat{\Delta}_{\calA}$   &$\hat{\Delta}_{\calX}$  \\
	\midrule
	$(\sfE_1,\sfE_2^{(1)})$ & $-0.002$ & $-0.002$ & $0.106$ & $0.002$ & $0.106$ \\
	$(\sfE_1,\sfE_2^{(2)})$ &  $-0.001$ & $-0.001$ & $0.086$ & $-0.012$ & $0.086$\\
    \bottomrule
    \end{tabular}
    }
    \caption{\textbf{Scores for \Cref{eg:app:1}.}} \label{tab:app:eg1}
\end{table}

\begin{table}
	\centering
	\resizebox{\columnwidth}{!}{%
	\begin{tabular}{lp{1.5cm}p{1.5cm}p{1.5cm}p{1.5cm}p{1.5cm}}
    \toprule
	\, & $\hat{\Delta}_1$& $\hat{\Delta}_2$ & $\hat{\Delta}_{\sfE_1}$ & $\hat{\Delta}_{\calA}$   &$\hat{\Delta}_{\calX}$  \\
	\midrule
	$(\sfE_1^{(1)},\sfE_2)$ &$-0.010$ & $-0.010$ & $0.068$ & $0.036$ & $0.068$ \\
	$(\sfE_1^{(2)},\sfE_2)$ &  $0.002$ & $0.001$ & $0.098$ & $0.035$ & $0.098$\\
    \bottomrule
    \end{tabular}
    }
    \caption{\textbf{Scores for \Cref{eg:app:2}.}} \label{tab:app:eg2}
\end{table}

\begin{table}
	\centering
	\resizebox{\columnwidth}{!}{%
	\begin{tabular}{lp{1.5cm}p{1.5cm}p{1.5cm}p{1.5cm}p{1.5cm}}
    \toprule
	\, & $\hat{\Delta}_1$& $\hat{\Delta}_2$ & $\hat{\Delta}_{\sfE_1}$ & $\hat{\Delta}_{\calA}$   &$\hat{\Delta}_{\calX}$  \\
	\midrule
	$(\sfE_1^{(1)},\sfE_2)$ & $0.056$ & $-0.001$ & $0.109$ & $0.096$ & $0.109$ \\
	$(\sfE_1^{(2)},\sfE_2)$ & $0.005$ & $-0.010$ & $0.279$ & $0.118$ & $0.279$\\
    \bottomrule
    \end{tabular}
    }
    \caption{\textbf{Scores for \Cref{eg:app:3}.}} \label{tab:app:eg3}
\end{table}

\subsubsection{Ablation Results}
    We report ablations on
    all possible subset of normalizations 
    together with temporality fine-tuning (-T,
    see \Cref{sec:inference} in \Cref{tab:app:ablation:full}.
    Note that when {\bf D} is enabled,
    {\bf S} and {\bf Q} are not active and
    when {\bf C} is enabled, {\bf E} is not active,
    thus resulting in a total of 
    $2^2(2^2+1)(2^1+1)=30$ combinations

Ablations resulting in the best performances
    are highlighted in blue and those resulting in the worst
    the performances are highlighted in red. Shaded rows
    are results without temporal fine-tuning (using top $k=30$
    tokens in mask language modeling).
    We summarize our observations as follows.
    
    \paragraph{Improvements due to normalizations are marginal.}
    The gap between best and worst performance
    are marginal, except for the \glt{} dataset, which
    is mainly caused by enabling
    estimand normalization ({\bf E}. Without considering {\bf E}, the worst result is $0.594$ ({\bf +Q} or {\bf +FQ}).
    Furthermore, we note the gap between
    the best results and the results under no normalizations
    ($\emptyset$) is also marginal, indicating that
    for CCR it is more important to have a well-established
    baseline and temporal signal extractors than exploring
    different normalizations.
    
    Furthermore, the outliers are interesting: enabling estimand
    normaliztion ({\bf E}) has little or no effects on
    most datasets but can boost the performance
    on \copate{} under non fine-tuned temporal predictors (-T)
    while is detrimental to \glt{} under fine-tuned temporal predictors.
    
    \paragraph{Rules-of-thumb for choosing normalizations.}
    As a general rule-of-thumb,
    temporal score normalization
	({\bf S}) should be enabled and the $q$ vectors should be properly formed
	(without direct matching {\bf D}); temporal pre-filtering ({\bf F}) and propensity normalization ({\bf Q})
	in general do not affect the results significantly;
	co-occurrence stabilization ({\bf C})
	has greater positive effect on datasets when a weaker notion of causation
	are desirable (e.g., \glt{} we constructed);
	while estimand normalization ({\bf E}) improves
	certain datasets (e.g., \copate{} without temporal fine-tuning), it has detrimental
	effects on some others (e.g., \glt{} with temporal fine-tuning),
	hence we recommend disabling it by default.

\subsection{Full Examples}
    We also attach three full examples from our
    implementation of the \ROCKY{}.
    The problem instances are given below.
    For each instance, we tabulate $50$ covariates
    sampled, all interventions generated, the
    corresponding $\norm{q(\x;\sfA)-q(\x;\sfE_1)}_p$,
    and the temporal probabilities $\Pr(\cdot \prec \sfE_2)$.
    
\begin{example}[Did $\sfE_1$ cause $\sfE_2^{(1)}$ or $\sfE_2^{(2)}$?] 
	{\small
	\begin{align*}
	    \sfE_1: &\quad \texttt{The teacher assigned homework to the students.} \\
	    \sfE_2^{(1)}: &\quad \texttt{The students passed notes.} \\
	    \sfE_2^{(2)}: &\quad \texttt{The students groaned.} \\
	\end{align*}
	}
	\label{eg:app:1}
\end{example}
    This is the $72$-nd instance of \copadev,
    the full tables for inferring the causation from
    $\sfE_1$ to $\sfE_2^{(1)}$
    and
    $\sfE_1$ to $\sfE_2^{(2)}$
    are given in \Cref{tab:app:smp:1a} and \Cref{tab:app:smp:1b}
    respectively. Different scores
    are shown in \Cref{tab:app:eg1}.
    Note that this example is not easy.

\begin{example}[Did $\sfE_1^{(1)}$
	or $\sfE_1^{(2)}$ cause $\sfE_2$?]
	{\small
	\begin{align*}
	    \sfE_1^{(1)}: &\quad \texttt{I was preparing to wash my hands.} \\
	    \sfE_1^{(2)}: &\quad \texttt{I was preparing to clean the bathroom.} \\
	    \sfE_2:&\quad \texttt{I put rubber gloves on.} \\
	\end{align*}
	} 
	\label{eg:app:2}
	\end{example}

    This is the $63$-rd instance of \copadev,
    the full tables for inferring the causation from
    $\sfE_1^{(1)}$ to $\sfE_2$
    and
    $\sfE_1^{(1)}$ to $\sfE_2^{(2)}$
    are given in \Cref{tab:app:smp:2a} and \Cref{tab:app:smp:2b}
    respectively.
    Different scores
    are shown in \Cref{tab:app:eg2}.

\begin{example}[Did $\sfE_1^{(1)}$
	or $\sfE_1^{(2)}$ cause $\sfE_2$?]
	{\small
	\begin{align*}
	    \sfE_1^{(1)}: &\quad \texttt{His pocket was filled with coins.} \\
	    \sfE_1^{(2)}: &\quad \texttt{He sewed the hole in his pocket.} \\
	    \sfE_2:&\quad \texttt{The man's pocket jingled as he walked.} \\
	\end{align*}
	} 
	\label{eg:app:3}
	\end{example}
	
    This is the $79$-th instance of \copadev,
    the full tables for inferring the causation from
    $\sfE_1^{(1)}$ to $\sfE_2$
    and
    $\sfE_1^{(1)}$ to $\sfE_2^{(2)}$
    are given in \Cref{tab:app:smp:3a} and \Cref{tab:app:smp:3b}
    respectively. Different scores
    are shown in \Cref{tab:app:eg1}.

\newcommand{\egtbhlt}{\cellcolor{Yellow!30}}
\clearpage
    
\begin{sidewaystable*}
    \centering
    \resizebox{\textwidth}{!}{%
	\begin{tabular}{ll|p{1cm}|p{1cm}|p{1cm}p{1cm}p{1cm}p{1cm}p{1cm}p{1cm}p{1cm}p{1cm}p{1cm}p{1cm}p{1cm}p{1cm}p{1cm}p{1cm}p{1cm}p{1cm}p{1cm}p{1cm}p{1cm}p{1cm}p{1cm}p{1cm}p{1cm}p{1cm}p{1cm}p{1cm}p{1cm}p{1cm}p{1cm}p{1cm}}
        \toprule
	Dataset & Score & Best & Worst &  $\emptyset$ & +\textbf{D} & +\textbf{F} & +\textbf{S} & +\textbf{Q} & +\textbf{C} & +\textbf{E} & +\textbf{D}\textbf{F} & +\textbf{D}\textbf{C} & +\textbf{D}\textbf{E} & +\textbf{F}\textbf{S} & +\textbf{F}\textbf{Q} & +\textbf{F}\textbf{C} & +\textbf{F}\textbf{E} & +\textbf{S}\textbf{Q} & +\textbf{S}\textbf{C} & +\textbf{S}\textbf{E} & +\textbf{Q}\textbf{C} & +\textbf{Q}\textbf{E} & +\textbf{D}\textbf{F}\textbf{C} & +\textbf{D}\textbf{F}\textbf{E} & +\textbf{F}\textbf{S}\textbf{Q} & +\textbf{F}\textbf{S}\textbf{Q}\textbf{C} & +\textbf{F}\textbf{S}\textbf{Q}\textbf{E} & +\textbf{F}\textbf{Q}\textbf{C} & +\textbf{F}\textbf{Q}\textbf{E} & +\textbf{S}\textbf{Q}\textbf{C} & +\textbf{S}\textbf{Q}\textbf{E} & +\textbf{F}\textbf{S}\textbf{Q}\textbf{C} & +\textbf{F}\textbf{S}\textbf{Q}\textbf{E}    \\
	\midrule
	\multirowcell{2}{\copadev} & $\hat{\Delta}_1$ $\uparrow$ &  $0.690$ & $0.620$ & $0.670$ & $0.660$ & $0.670$ & $0.680$ & $0.650$ & $0.650$ & $0.680$ & $0.660$ & $0.650$ & $0.680$ & $0.680$ & $0.650$ & $0.650$ & $0.680$ & $0.670$ & \abpoor{$0.620$} & $0.680$ & $0.640$ & $0.660$ & $0.650$ & $0.680$ & $0.670$ & $0.660$ & \abgood{$0.690$} & $0.640$ & $0.660$ & $0.660$ & \abgood{$0.690$} & $0.660$ & \abgood{$0.690$}  \\
	\, & $\hat{\Delta}_2$ $\uparrow$ & $0.700$ & $0.630$ & \abpoor{$0.630$} & $0.650$ & \abpoor{$0.630$} & $0.690$ & \abpoor{$0.630$} & $0.660$ & $0.640$ & $0.650$ & $0.650$ & $0.680$ & $0.690$ & \abpoor{$0.630$} & $0.660$ & $0.640$ & $0.670$ & \abpoor{$0.630$} & \abgood{$0.700$} & $0.650$ & $0.660$ & $0.650$ & $0.680$ & $0.670$ & $0.640$ & \abgood{$0.700$} & $0.650$ & $0.660$ & $0.640$ & \abgood{$0.700$} & $0.640$ & \abgood{$0.700$}   \\
	\multirowcell{2}{\copate} &  $\hat{\Delta}_1$ $\uparrow$  & $0.564$ & $0.528$ & $0.542$ & $0.548$ & $0.542$ & $0.540$ & $0.548$ & \abgood{$0.564$} & $0.554$ & $0.548$ & \abgood{$0.564$} & $0.554$ & $0.540$ & $0.548$ & \abgood{$0.564$} & $0.554$ & $0.532$ & \abgood{$0.564$} & $0.532$ & $0.558$ & $0.560$ & \abgood{$0.564$} & $0.554$ & $0.532$ & $0.560$ & \abpoor{$0.528$} & $0.558$ & $0.560$ & $0.560$ & \abpoor{$0.528$} & $0.560$ & \abpoor{$0.528$}  \\
	\, & $\hat{\Delta}_2$ $\uparrow$ & $0.564$ & $0.526$ & $0.554$ & $0.542$ & $0.554$ & $0.540$ & $0.544$ & \abgood{$0.564$} & $0.548$ & $0.542$ & \abgood{$0.564$} & $0.546$ & $0.540$ & $0.544$ & \abgood{$0.564$} & $0.548$ & $0.538$ & \abgood{$0.564$} & $0.534$ & $0.562$ & $0.556$ & \abgood{$0.564$} & $0.546$ & $0.538$ & $0.562$ & \abpoor{$0.526$} & $0.562$ & $0.556$ & $0.562$ & \abpoor{$0.526$} & $0.562$ & \abpoor{$0.526$} \\
	\multirowcell{2}{\glt} &$\hat{\Delta}_1$ $\uparrow$& $0.665$ & $0.503$ & $0.600$ & $0.606$ & $0.600$ & $0.594$ & $0.594$ & $0.613$ & \abpoor{$0.503$} & $0.606$ & $0.639$ & \abpoor{$0.503$} & $0.594$ & $0.594$ & $0.613$ & \abpoor{$0.503$} & $0.594$ & $0.639$ & $0.510$ & $0.613$ & $0.510$ & $0.639$ & \abpoor{$0.503$} & $0.594$ & \abgood{$0.665$} & $0.516$ & $0.613$ & $0.510$ & \abgood{$0.665$} & $0.516$ & \abgood{$0.665$} & $0.516$  \\
	\, & $\hat{\Delta}_2$ $\uparrow$ & $0.697$ & $0.503$ & $0.594$ & $0.600$ & $0.594$ & $0.606$ & $0.594$ & $0.619$ & $0.510$ & $0.600$ & $0.639$ & \abpoor{$0.503$} & $0.606$ & $0.594$ & $0.619$ & $0.510$ & $0.600$ & \abgood{$0.697$} & $0.510$ & $0.619$ & $0.516$ & $0.639$ & \abpoor{$0.503$} & $0.600$ & $0.690$ & $0.516$ & $0.619$ & $0.516$ & $0.690$ & $0.516$ & $0.690$ & $0.516$  \\
	\midrule 
	\belowrulesepcolor{Gray!20}
	\rowcolor{Gray!20}
	\, & $\hat{\Delta}_1$ $\uparrow$ &  $0.620$ & $0.550$ & $0.590$ & \abpoor{$0.550$} & $0.590$ & $0.580$ & $0.580$ & $0.570$ & \abgood{$0.620$} & \abpoor{$0.550$} & $0.560$ & $0.610$ & $0.580$ & $0.580$ & $0.570$ & \abgood{$0.620$} & $0.580$ & $0.560$ & \abgood{$0.620$} & $0.560$ & \abgood{$0.620$} & $0.560$ & $0.610$ & $0.580$ & $0.560$ & $0.610$ & $0.560$ & \abgood{$0.620$} & $0.560$ & $0.610$ & $0.560$ & $0.610$  \\
	\rowcolor{Gray!20}
	\multirow{-2}{*}{\copadev{} (-T)} & $\hat{\Delta}_2$ $\uparrow$ & $0.630$ & $0.530$ & $0.610$ & \abpoor{$0.530$} & $0.610$ & $0.600$ & $0.580$ & $0.600$ & \abgood{$0.630$} & \abpoor{$0.530$} & $0.550$ & $0.600$ & $0.600$ & $0.580$ & $0.600$ & \abgood{$0.630$} & $0.580$ & $0.580$ & $0.610$ & $0.580$ & $0.620$ & $0.550$ & $0.600$ & $0.580$ & $0.560$ & $0.610$ & $0.580$ & $0.620$ & $0.560$ & $0.610$ & $0.560$ & $0.610$  \\
	\rowcolor{Gray!20}
	\, &  $\hat{\Delta}_1$ $\uparrow$  &$0.580$ & $0.484$ & $0.494$ & $0.486$ & $0.494$ & $0.522$ & $0.498$ & \abpoor{$0.484$} & $0.574$ & $0.486$ & $0.496$ & $0.574$ & $0.522$ & $0.498$ & \abpoor{$0.484$} & $0.574$ & $0.514$ & $0.512$ & $0.570$ & $0.506$ & \abgood{$0.580$} & $0.496$ & $0.574$ & $0.514$ & $0.512$ & $0.570$ & $0.506$ & \abgood{$0.580$} & $0.512$ & $0.570$ & $0.512$ & $0.570$   \\
	\rowcolor{Gray!20}
	\multirow{-2}{*}{\copate{} (-T)} & $\hat{\Delta}_2$ $\uparrow$ & $0.574$ & $0.484$ & $0.494$ & $0.502$ & $0.494$ & $0.530$ & $0.508$ & \abpoor{$0.484$} & \abgood{$0.574$} & $0.502$ & $0.492$ & $0.570$ & $0.530$ & $0.508$ & \abpoor{$0.484$} & \abgood{$0.574$} & $0.528$ & $0.524$ & $0.570$ & $0.494$ & \abgood{$0.574$} & $0.492$ & $0.570$ & $0.528$ & $0.522$ & $0.570$ & $0.494$ & \abgood{$0.574$} & $0.522$ & $0.570$ & $0.522$ & $0.570$   \\
	\rowcolor{Gray!20}
	\, &$\hat{\Delta}_1$ $\uparrow$& $0.606$ & $0.510$ & $0.568$ & $0.555$ & $0.568$ & $0.574$ & $0.568$ & $0.535$ & \abgood{$0.606$} & $0.555$ & \abpoor{$0.510$} & $0.587$ & $0.574$ & $0.568$ & $0.535$ & \abgood{$0.606$} & $0.568$ & $0.529$ & \abgood{$0.606$} & $0.542$ & $0.594$ & \abpoor{$0.510$} & $0.587$ & $0.568$ & $0.535$ & $0.600$ & $0.542$ & $0.594$ & $0.535$ & $0.600$ & $0.535$ & $0.600$   \\
	\rowcolor{Gray!20}
	\multirowcell{-2}{\glt{} (-T)} & $\hat{\Delta}_2$ $\uparrow$ & $0.619$ & $0.503$ & $0.568$ & $0.555$ & $0.568$ & $0.587$ & $0.561$ & \abpoor{$0.503$} & \abgood{$0.619$} & $0.555$ & $0.516$ & $0.587$ & $0.587$ & $0.561$ & \abpoor{$0.503$} & \abgood{$0.619$} & $0.581$ & $0.529$ & $0.613$ & $0.548$ & $0.594$ & $0.516$ & $0.587$ & $0.581$ & $0.535$ & $0.613$ & $0.548$ & $0.594$ & $0.535$ & $0.613$ & $0.535$ & $0.613$  \\
	\aboverulesepcolor{Gray!20}
    \bottomrule
    \end{tabular}
    }
    \caption{\textbf{Full ablation studies on normalizations.} Ablations resulting in the best performances
    are highlighted in blue and those resulting in the worst
    performances are highlighted in red. Shaded rows
    are results without temporal fine-tuning (using top $k=30$
    tokens in masked language modeling).
    (i) The gaps between best and worst performance
    are marginal, except for the \glt{} dataset, which
    is mainly due to
    estimand normalization {\bf E}. Without considering {\bf E}, the worst result is $0.594$ ({\bf +Q} or {\bf +FQ}).
    (ii) In general, temporal fine-tuning helps. The only
    exception on \copate{} is due to
    estimand normalization ({\bf E}).
    (iii) As a general rule-of-thumb,
    it does not hurt to start with no normalizations
    enabled.
	}
    \label{tab:app:ablation:full}
\end{sidewaystable*}

\clearpage
    \begin{sidewaystable*}
    \centering
    \resizebox{\textwidth}{!}{%
	\begin{tabular}{p{15cm}p{4cm}p{28cm}p{4cm}}
        \toprule
	Sampled Covariates $\calX$ & $\norm{q(\x;\sfA)-q(\x;\sfE_1)}_p$ & $\sfE_1$ and Interventions $\calA$ & $\Pr(\cdot \prec \sfE_2)$\\
    \midrule
    &  \multirowcell{40}[0em][l]{\, \\
$0$ \\
$0.0135$ \\
$0.0508$ \\
$0.0894$ \\
$0.0279$ \\
$0.1053$ \\
$0.1291$ \\
$0.0591$ \\
$0.0365$ \\
$0.0201$ \\
$0.0870$ \\
$0.1521$ \\
$0.0820$ \\
$0.1524$ \\
$0.1349$ \\
$0.1999$ \\
$0.0468$ \\
$0.0485$ \\
$0.0362$ \\
$0.0301$ \\
$0.0135$ \\
$0.0488$ \\
$0.0512$ \\
$0.0515$ \\
$0.0201$ \\
$0.0647$ \\
$0.0298$ \\} & \multirowcell{40}[0em][l]{\, \\
{\small $\sfE_1$: \texttt{The teacher assigned homework to the students.}} \\
\egtbhlt {\small $\sfA_{1}$: \texttt{The professor assigned homework to the students.}} \\
{\small $\sfA_{2}$: \texttt{The professor supported the tourists assigned homework to the students.}} \\
{\small $\sfA_{3}$: \texttt{The tourists ran, or the teacher assigned homework to the students.}} \\
\egtbhlt {\small $\sfA_{4}$: \texttt{The teacher took homework to the students.}} \\
{\small $\sfA_{5}$: \texttt{The teacher was assigning Justin with the homework to the students.}} \\
{\small $\sfA_{6}$: \texttt{The teacher replaced the carpet for the library last night because the carpet was old homework to the students.}} \\
{\small $\sfA_{7}$: \texttt{The teacher assigned to read the children's book came to the students.}} \\
\egtbhlt {\small $\sfA_{8}$: \texttt{The teacher assigned less homework to the students.}} \\
\egtbhlt {\small $\sfA_{9}$: \texttt{The teacher assigned tests to the students.}} \\
{\small $\sfA_{10}$: \texttt{No one was assigned homework to the students.}} \\
{\small $\sfA_{11}$: \texttt{Unless the senator performed, the teacher assigned homework to the students.}} \\
{\small $\sfA_{12}$: \texttt{Noelle Leong on the other hand assigned homework to the students.}} \\
{\small $\sfA_{13}$: \texttt{The teacher didn't give homework to the students.}} \\
{\small $\sfA_{14}$: \texttt{The teacher didn't assigned homework to the students.}} \\
{\small $\sfA_{15}$: \texttt{The teacher didn't tell anyone homework to the students.}} \\
{\small $\sfA_{16}$: \texttt{The teacher assigned nothing to the students.}} \\
{\small $\sfA_{17}$: \texttt{The teacher assigned no children to the students.}} \\
\egtbhlt {\small $\sfA_{18}$: \texttt{The teacher assigned no class to the students.}} \\
\egtbhlt {\small $\sfA_{19}$: \texttt{The student assigned homework to the students.}} \\
\egtbhlt {\small $\sfA_{20}$: \texttt{The professor assigned homework to the students.}} \\
{\small $\sfA_{21}$: \texttt{The teacher worked on the algebraic homework to the students.}} \\
{\small $\sfA_{22}$: \texttt{The teacher wrote homework to the students.}} \\
{\small $\sfA_{23}$: \texttt{The teacher read homework to the students.}} \\
\egtbhlt {\small $\sfA_{24}$: \texttt{The teacher assigned tests to the students.}} \\
{\small $\sfA_{25}$: \texttt{The teacher assigned to the classroom stopped to the students.}} \\
\egtbhlt {\small $\sfA_{26}$: \texttt{The teacher assigned anger to the students.}} \\} & \multirowcell{40}[0em][l]{\, \\
$0.5031$ \\
$0.4993$ \\
$0.5082$ \\
$0.4987$ \\
$0.5177$ \\
$0.3935$ \\
$0.5093$ \\
$0.5120$ \\
$0.5135$ \\
$0.4999$ \\
$0.4886$ \\
$0.3867$ \\
$0.4874$ \\
$0.5324$ \\
$0.5198$ \\
$0.5140$ \\
$0.5175$ \\
$0.5230$ \\
$0.5167$ \\
$0.4958$ \\
$0.4993$ \\
$0.5334$ \\
$0.5151$ \\
$0.5231$ \\
$0.4999$ \\
$0.5244$ \\
$0.5065$ \\} \\
  \partialinput{1}{40}{tables/raw_data/eg_copa_dev_72_1_cov.txt}
  \, & \, & \, & \, \\
    \bottomrule
    \end{tabular}
    }
    \caption{\textbf{Example 1a:} the first plausible
    pair of the $72$-th instance
    in \copadev, matched interventions are highlighted.
    Here $\sfE_1:$ \texttt{The teacher assigned homework to the students.} and $\sfE_2$: \texttt{The students passed notes.}
    }
    \label{tab:app:smp:1a}
\end{sidewaystable*}

\clearpage
    \begin{sidewaystable*}
    \centering
    \resizebox{\textwidth}{!}{%
	\begin{tabular}{p{15cm}p{4cm}p{25cm}p{4cm}}
        \toprule
	Sampled Covariates $\calX$ & $\norm{q(\x;\sfA)-q(\x;\sfE_1)}_p$ & $\sfE_1$ and Interventions $\calA$ & $\Pr(\cdot \prec \sfE_2)$\\
    \midrule
    &  \multirowcell{40}[0em][l]{\, \\
$0$ \\
$0.0135$ \\
$0.0508$ \\
$0.0894$ \\
$0.0279$ \\
$0.1053$ \\
$0.1291$ \\
$0.0591$ \\
$0.0365$ \\
$0.0201$ \\
$0.0870$ \\
$0.1521$ \\
$0.0820$ \\
$0.1524$ \\
$0.1349$ \\
$0.1999$ \\
$0.0468$ \\
$0.0485$ \\
$0.0362$ \\
$0.0301$ \\
$0.0135$ \\
$0.0488$ \\
$0.0512$ \\
$0.0515$ \\
$0.0201$ \\
$0.0647$ \\
$0.0298$ \\} & \multirowcell{40}[0em][l]{\, \\
{\small $\sfE_1$: \texttt{The teacher assigned homework to the students.}} \\
\egtbhlt {\small $\sfA_{1}$: \texttt{The professor assigned homework to the students.}} \\
{\small $\sfA_{2}$: \texttt{The professor supported the tourists assigned homework to the students.}} \\
{\small $\sfA_{3}$: \texttt{The tourists ran, or the teacher assigned homework to the students.}} \\
\egtbhlt {\small $\sfA_{4}$: \texttt{The teacher took homework to the students.}} \\
{\small $\sfA_{5}$: \texttt{The teacher was assigning Justin with the homework to the students.}} \\
{\small $\sfA_{6}$: \texttt{The teacher replaced the carpet for the library last night because the carpet was old homework to the students.}} \\
{\small $\sfA_{7}$: \texttt{The teacher assigned to read the children's book came to the students.}} \\
\egtbhlt {\small $\sfA_{8}$: \texttt{The teacher assigned less homework to the students.}} \\
\egtbhlt {\small $\sfA_{9}$: \texttt{The teacher assigned tests to the students.}} \\
{\small $\sfA_{10}$: \texttt{No one was assigned homework to the students.}} \\
{\small $\sfA_{11}$: \texttt{Unless the senator performed, the teacher assigned homework to the students.}} \\
{\small $\sfA_{12}$: \texttt{Noelle Leong on the other hand assigned homework to the students.}} \\
{\small $\sfA_{13}$: \texttt{The teacher didn't give homework to the students.}} \\
{\small $\sfA_{14}$: \texttt{The teacher didn't assigned homework to the students.}} \\
{\small $\sfA_{15}$: \texttt{The teacher didn't tell anyone homework to the students.}} \\
{\small $\sfA_{16}$: \texttt{The teacher assigned nothing to the students.}} \\
{\small $\sfA_{17}$: \texttt{The teacher assigned no children to the students.}} \\
\egtbhlt {\small $\sfA_{18}$: \texttt{The teacher assigned no class to the students.}} \\
\egtbhlt {\small $\sfA_{19}$: \texttt{The student assigned homework to the students.}} \\
\egtbhlt {\small $\sfA_{20}$: \texttt{The professor assigned homework to the students.}} \\
{\small $\sfA_{21}$: \texttt{The teacher worked on the algebraic homework to the students.}} \\
{\small $\sfA_{22}$: \texttt{The teacher wrote homework to the students.}} \\
{\small $\sfA_{23}$: \texttt{The teacher read homework to the students.}} \\
\egtbhlt {\small $\sfA_{24}$: \texttt{The teacher assigned tests to the students.}} \\
{\small $\sfA_{25}$: \texttt{The teacher assigned to the classroom stopped to the students.}} \\
\egtbhlt {\small $\sfA_{26}$: \texttt{The teacher assigned anger to the students.}} \\} & \multirowcell{40}[0em][l]{\, \\
$0.5308$ \\
$0.5263$ \\
$0.5207$ \\
$0.5260$ \\
$0.5340$ \\
$0.5396$ \\
$0.5015$ \\
$0.5346$ \\
$0.5515$ \\
$0.5249$ \\
$0.5832$ \\
$0.5454$ \\
$0.5288$ \\
$0.5914$ \\
$0.5956$ \\
$0.6164$ \\
$0.5487$ \\
$0.5566$ \\
$0.5477$ \\
$0.5180$ \\
$0.5263$ \\
$0.5349$ \\
$0.5318$ \\
$0.5370$ \\
$0.5249$ \\
$0.5263$ \\
$0.5291$ \\} \\
\partialinput{1}{40}{tables/raw_data/eg_copa_dev_72_2_cov.txt}
    \, & \, & \, & \, \\
    \bottomrule
    \end{tabular}
    }
    \caption{\textbf{Example 1b:} the second plausible
    pair of the $72$-th instance
    in \copadev, matched interventions are highlighted.
    Here $\sfE_1:$ \texttt{The teacher assigned homework to the students.} and $\sfE_2$:  \texttt{The students groaned.}
    }
    \label{tab:app:smp:1b}
\end{sidewaystable*}

\clearpage
    \begin{sidewaystable*}
    \centering
    \resizebox{\textwidth}{!}{%
	\begin{tabular}{p{15cm}p{4cm}p{28cm}p{4cm}}
        \toprule
	Sampled Covariates $\calX$ & $\norm{q(\x;\sfA)-q(\x;\sfE_1)}_p$ & $\sfE_1$ and Interventions $\calA$ & $\Pr(\cdot \prec \sfE_2)$\\
    \midrule
    &  \multirowcell{50}[0em][l]{\partialinput{1}{50}{tables/raw_data/eg_copa_dev_63_1_dist.txt}} & \multirowcell{50}[0em][l]{\partialinput{1}{50}{tables/raw_data/eg_copa_dev_63_1_d.txt}} & \multirowcell{50}[0em][l]{\partialinput{1}{50}{tables/raw_data/eg_copa_dev_63_1_pdy.txt}} \\
  \partialinput{1}{50}{tables/raw_data/eg_copa_dev_63_1_cov.txt}
  \, & \, & \, & \, \\
    \bottomrule
    \end{tabular}
    }
    \caption{\textbf{Example 2a:} the first plausible
    pair of the $63$-th instance
    in \copadev, matched interventions are highlighted.
    Here $\sfE_1:$ \texttt{I was preparing to wash my hands.} and $\sfE_2$:  \texttt{I put rubber gloves on.}
    }
    \label{tab:app:smp:2a}
\end{sidewaystable*}

\clearpage
    \begin{sidewaystable*}
    \centering
    \resizebox{\textwidth}{!}{%
	\begin{tabular}{p{15cm}p{4cm}p{25cm}p{4cm}}
        \toprule
	Sampled Covariates $\calX$ & $\norm{q(\x;\sfA)-q(\x;\sfE_1)}_p$ & $\sfE_1$ and Interventions $\calA$ & $\Pr(\cdot \prec \sfE_2)$\\
    \midrule
    &  \multirowcell{50}[0em][l]{\partialinput{1}{50}{tables/raw_data/eg_copa_dev_63_2_dist.txt}} & \multirowcell{50}[0em][l]{\partialinput{1}{50}{tables/raw_data/eg_copa_dev_63_2_d.txt}} & \multirowcell{50}[0em][l]{\partialinput{1}{50}{tables/raw_data/eg_copa_dev_63_2_pdy.txt}} \\
\partialinput{1}{50}{tables/raw_data/eg_copa_dev_63_2_cov.txt}
\, & \, & \, & \, \\
    \bottomrule
    \end{tabular}
    }
    \caption{\textbf{Example 2b:} the second plausible
    pair of the $63$-th instance
    in \copadev, matched interventions are highlighted.
    Here $\sfE_1:$ \texttt{I was preparing to clean the bathroom.} and $\sfE_2$:  \texttt{I put rubber gloves on.}
    }
    \label{tab:app:smp:2b}
\end{sidewaystable*}

\clearpage
    \begin{sidewaystable*}
    \centering
    \resizebox{\textwidth}{!}{%
	\begin{tabular}{p{15cm}p{4cm}p{28cm}p{4cm}}
        \toprule
	Sampled Covariates $\calX$ & $\norm{q(\x;\sfA)-q(\x;\sfE_1)}_p$ & $\sfE_1$ and Interventions $\calA$ & $\Pr(\cdot \prec \sfE_2)$\\
    \midrule
    &  \multirowcell{40}[0em][l]{\, \\
$0$ \\
$0.1069$ \\
$0.0706$ \\
$0.0620$ \\
$0.1457$ \\
$0.0626$ \\
$0.0938$ \\
$0.0747$ \\
$0.0364$ \\
$0.1158$ \\
$0.2276$ \\
$0.1331$ \\
$0.1170$ \\
$0.1539$ \\
$0.1590$ \\
$0.1126$ \\
$0.3496$ \\
$0.1112$ \\
$0.0897$ \\
$0.0000$ \\
$0.0820$ \\
$0.0239$ \\
$0.2621$ \\
$0.0233$ \\
$0.0481$ \\
$0.0753$ \\
$0.0370$ \\} & \multirowcell{40}[0em][l]{\, \\
{\small $\sfE_1$: \texttt{His pocket was filled with coins.}} \\
{\small $\sfA_{1}$: \texttt{His pocket however had been filled with coins.}} \\
{\small $\sfA_{2}$: \texttt{His pocket contained nine corsage filled with coins.}} \\
{\small $\sfA_{3}$: \texttt{His pocket had a large amount of space filled with coins.}} \\
{\small $\sfA_{4}$: \texttt{His pocket was lost when he accidentally took some with coins.}} \\
{\small $\sfA_{5}$: \texttt{His pocket was empty with coins.}} \\
{\small $\sfA_{6}$: \texttt{His pocket was filled with sandals and at least one with coins.}} \\
{\small $\sfA_{7}$: \texttt{A cowboy on the back of a wagon was filled with coins.}} \\
\egtbhlt {\small $\sfA_{8}$: \texttt{A pocket iron was filled with coins.}} \\
{\small $\sfA_{9}$: \texttt{Mark was filled with coins.}} \\
{\small $\sfA_{10}$: \texttt{His pocket did not work as well as the shirt, because the shirt was filled with coins.}} \\
{\small $\sfA_{11}$: \texttt{His pocket wasn't filled with coins.}} \\
{\small $\sfA_{12}$: \texttt{His pocket was not filled with coins.}} \\
{\small $\sfA_{13}$: \texttt{His pocket was empty but his pocket had nine with coins.}} \\
{\small $\sfA_{14}$: \texttt{His pocket was not holding the lotion with coins.}} \\
{\small $\sfA_{15}$: \texttt{His pocket was not touched with coins.}} \\
{\small $\sfA_{16}$: \texttt{No matter how you feel about country music [...] the fact that it featured the really catchy John Denver does not appeal to me was filled with coins.}} \\
{\small $\sfA_{17}$: \texttt{No pocket book was filled with coins.}} \\
{\small $\sfA_{18}$: \texttt{Nothing was filled with coins.}} \\
{\small $\sfA_{19}$: \texttt{His pocket was filled with coins.}} \\
{\small $\sfA_{20}$: \texttt{His pocket had been filled with coins.}} \\
\egtbhlt {\small $\sfA_{21}$: \texttt{His pocket was heavily filled with coins.}} \\
{\small $\sfA_{22}$: \texttt{His pocket was ruined and he decided to find a wallet instead because the pocket might have coins with coins.}} \\
\egtbhlt {\small $\sfA_{23}$: \texttt{His pocket was full with coins.}} \\
{\small $\sfA_{24}$: \texttt{Her bag was filled with coins.}} \\
{\small $\sfA_{25}$: \texttt{Someone was filled with coins.}} \\
\egtbhlt {\small $\sfA_{26}$: \texttt{Her wallet was filled with coins.}} \\} & \multirowcell{40}[0em][l]{\, \\
$0.2980$ \\
$0.1494$ \\
$0.3912$ \\
$0.3036$ \\
$0.0620$ \\
$0.2749$ \\
$0.2597$ \\
$0.2331$ \\
$0.1933$ \\
$0.1925$ \\
$0.0994$ \\
$0.1870$ \\
$0.2477$ \\
$0.1345$ \\
$0.0834$ \\
$0.1852$ \\
$0.0209$ \\
$0.1971$ \\
$0.2345$ \\
$0.2980$ \\
$0.3352$ \\
$0.2535$ \\
$0.0913$ \\
$0.3068$ \\
$0.1762$ \\
$0.2306$ \\
$0.2127$ \\} \\
  \partialinput{1}{40}{tables/raw_data/eg_copa_dev_79_1_cov.txt}
  \, & \, & \, & \, \\
    \bottomrule
    \end{tabular}
    }
    \caption{\textbf{Example 3a:} the first plausible
    pair of the $79$-th instance
    in \copadev, matched interventions are highlighted.
    Here $\sfE_1:$ \texttt{His pocket was filled with coins.} and $\sfE_2$:  \texttt{The man's pocket jingled as he walked.}
    }
    \label{tab:app:smp:3a}
\end{sidewaystable*}

\clearpage
    \begin{sidewaystable*}
    \centering
    \resizebox{\textwidth}{!}{%
	\begin{tabular}{p{15cm}p{4cm}p{25cm}p{4cm}}
        \toprule
	Sampled Covariates $\calX$ & $\norm{q(\x;\sfA)-q(\x;\sfE_1)}_p$ & $\sfE_1$ and Interventions $\calA$ & $\Pr(\cdot \prec \sfE_2)$\\
    \midrule
    &  \multirowcell{40}[0em][l]{\, \\
$0$ \\
$0.1075$ \\
$0.2456$ \\
$0.0682$ \\
$0.0660$ \\
$0.0875$ \\
$0.0849$ \\
$0.2149$ \\
$0.0930$ \\
$0.0707$ \\
$0.1181$ \\
$0.2376$ \\
$0.1161$ \\
$0.1128$ \\
$0.1479$ \\
$0.0869$ \\
$0.1401$ \\
$0.3286$ \\
$0.0871$ \\
$0.0764$ \\
$0.0456$ \\
$0.0715$ \\
$0.1560$ \\
$0.0614$ \\
$0.0555$ \\
$0.0588$ \\
$0.0311$ \\
$0.0499$ \\} & \multirowcell{40}[0em][l]{\, \\
{\small $\sfE_1$: \texttt{He sewed the hole in his pocket.}} \\
{\small $\sfA_{1}$: \texttt{Then he sewed the hole in his pocket.}} \\
{\small $\sfA_{2}$: \texttt{The boy was grumpy in high school, but happy at school, so the teacher taught him sewed the hole in his pocket.}} \\
{\small $\sfA_{3}$: \texttt{He cut pieces from the plate sewed the hole in his pocket.}} \\
{\small $\sfA_{4}$: \texttt{He stuffed the toy gun with the hole in his pocket.}} \\
{\small $\sfA_{5}$: \texttt{He burnt the hole by pulling the hole in his pocket.}} \\
{\small $\sfA_{6}$: \texttt{He pulled off the blanket and got a the hole in his pocket.}} \\
{\small $\sfA_{7}$: \texttt{He sewed the quilt better than a teepee because the teepee was a sloppy job in his pocket.}} \\
{\small $\sfA_{8}$: \texttt{He sewed with a towel more in his pocket.}} \\
{\small $\sfA_{9}$: \texttt{He sewed the hole with a wire instead of a plier in his pocket.}} \\
{\small $\sfA_{10}$: \texttt{He couldn't sewed the hole in his pocket.}} \\
{\small $\sfA_{11}$: \texttt{No matter how you feel about country music( I for one can't stand it despite my Houston roots), this only instilled sewed the hole in his pocket.}} \\
{\small $\sfA_{12}$: \texttt{No one knew how to sewed the hole in his pocket.}} \\
{\small $\sfA_{13}$: \texttt{He never filled the hole in his pocket.}} \\
{\small $\sfA_{14}$: \texttt{He couldn't bend the iron rod and instead tied the hole in his pocket.}} \\
{\small $\sfA_{15}$: \texttt{He had the hole in his pocket.}} \\
{\small $\sfA_{16}$: \texttt{He sewed no better than the tleilaxu which cut off his eye in his pocket.}} \\
{\small $\sfA_{17}$: \texttt{He sewed no better with the machine than with the method, because the machine was not precise in his pocket.}} \\
{\small $\sfA_{18}$: \texttt{He sewed not only the hole but also the whole ball inside the hole in his pocket.}} \\
{\small $\sfA_{19}$: \texttt{Jack sewed the hole in his pocket.}} \\
{\small $\sfA_{20}$: \texttt{Someone sewed the hole in his pocket.}} \\
{\small $\sfA_{21}$: \texttt{He screwed up sewed the hole in his pocket.}} \\
{\small $\sfA_{22}$: \texttt{He flunked out of high school, ended up in a strange town, and started writing about the weird the hole in his pocket.}} \\
{\small $\sfA_{23}$: \texttt{He stabbed Wisbech with a mop the hole in his pocket.}} \\
{\small $\sfA_{24}$: \texttt{He pulled the hole in his pocket.}} \\
{\small $\sfA_{25}$: \texttt{He sewed the turkey with a T-shirt in his pocket.}} \\
\egtbhlt {\small $\sfA_{26}$: \texttt{He sewed the bell necklace in his pocket.}} \\
{\small $\sfA_{27}$: \texttt{He sewed the rope with a chisel in his pocket.}} \\} & \multirowcell{40}[0em][l]{\, \\
$0.4818$ \\
$0.3724$ \\
$0.1477$ \\
$0.5023$ \\
$0.5053$ \\
$0.3826$ \\
$0.4970$ \\
$0.2377$ \\
$0.3728$ \\
$0.4299$ \\
$0.3049$ \\
$0.1855$ \\
$0.2986$ \\
$0.2366$ \\
$0.2361$ \\
$0.3282$ \\
$0.1917$ \\
$0.0641$ \\
$0.4182$ \\
$0.4594$ \\
$0.4086$ \\
$0.4874$ \\
$0.3031$ \\
$0.4916$ \\
$0.4673$ \\
$0.5179$ \\
$0.4765$ \\
$0.4956$ \\} \\
\partialinput{1}{40}{tables/raw_data/eg_copa_dev_79_2_cov.txt}
\, & \, & \, & \, \\
    \bottomrule
    \end{tabular}
    }
    \caption{\textbf{Example 3b:} the second plausible
    pair of the $79$-th instance
    in \copadev, matched interventions are highlighted.
    Here $\sfE_1:$ \texttt{He sewed the hole in his pocket.} and $\sfE_2$:  \texttt{The man's pocket jingled as he walked.}
    }
    \label{tab:app:smp:3b}
\end{sidewaystable*}



\end{document}